\documentclass{article}

\usepackage{arxiv}

\usepackage[utf8]{inputenc} 
\usepackage[T1]{fontenc}    
\usepackage{hyperref}       
\usepackage{url}            
\usepackage{booktabs}       
\usepackage{amsfonts}       
\usepackage{nicefrac}       
\usepackage{microtype}      
\usepackage{lipsum}

\usepackage{relsize}
\usepackage{verbatim}
\usepackage{graphicx}
\usepackage{dcolumn}
\usepackage{bm}
\usepackage{float}

\usepackage{graphics}
\usepackage{color}
\usepackage{xcolor}
\usepackage{bm}
\usepackage{subfigure}

\usepackage{soul}
\usepackage{xcolor}
\usepackage{graphicx}
\usepackage{hyperref}
\usepackage{subcaption}   
\usepackage{subcaption}
\usepackage{tikz}
\usetikzlibrary{arrows.meta,positioning,fit,calc}
\usepackage{booktabs}
\usepackage{bm}

\usepackage{amsmath,amssymb,amsfonts}

\usepackage{algorithm}
\usepackage{algpseudocode}
\algnewcommand\server{\item[\textbf{Server execution:}]}%
\algnewcommand\client{\item[\textbf{ClientUpdate($k,w$):}]}%

\usepackage{enumitem}
\setlist[itemize]{leftmargin=*}
\setlist[enumerate]{leftmargin=*}
\usepackage{amsmath,amssymb,amsfonts}
\usepackage{graphics}
\usepackage{color}
\usepackage{xcolor}
\definecolor{rev}{rgb}{0,0,0}
\definecolor{rev2}{rgb}{0,0,0}

\usepackage{array}
\newcolumntype{P}[1]{>{\centering\arraybackslash}p{#1}}
\usepackage{multirow}
\usepackage{cancel}
\usepackage[capitalise]{cleveref} 
\usepackage{enumitem}

\usepackage{cite}

\usepackage{tabularx}
\usepackage{threeparttable}
\usepackage{pifont} 
\newcolumntype{Y}{>{\centering\arraybackslash}X}

\title{Spectral Embedding via Chebyshev Bases for Robust DeepONet Approximation}

\author{
  Muhammad Abid \\
  Department of Mechanical and Aerospace Engineering,\\
  University of Tennessee,\\
  Knoxville, TN 37996, USA.\\
  \texttt{mabid@vols.utk.edu} 
  \And
  Omer San \\
  Department of Mechanical and Aerospace Engineering,\\
  University of Tennessee,\\
  Knoxville, TN 37996, USA.\\
  \texttt{osan@utk.edu}
}

\begin{document}
\maketitle

\begin{abstract}
Deep Operator Networks (DeepONets) have become a central tool in data-driven operator learning, providing flexible surrogates for nonlinear mappings arising in partial differential equations (PDEs). However, the standard trunk design based on fully connected layers acting on raw spatial or spatiotemporal coordinates struggles to represent sharp gradients, boundary layers, and other non-periodic structures commonly found in PDEs posed on bounded domains with Dirichlet or Neumann boundary conditions. To address these limitations, we introduce the Spectral-Embedded DeepONet (SEDONet), a new DeepONet variant in which the trunk is driven by a fixed Chebyshev spectral dictionary rather than coordinate inputs. This non-periodic spectral embedding provides a principled inductive bias tailored to bounded domains, enabling the learned operator to capture fine-scale non-periodic features that are difficult for Fourier-based or MLP-only trunks to represent. SEDONet is evaluated on a suite of benchmark problems including the 2-D Poisson equation, 1-D Burgers' equation, 1-D advection-diffusion equation, Allen-Cahn equation, Lorenz-96 chaotic system, and Darcy flow, covering elliptic, hyperbolic, parabolic, chaotic, and multiscale phenomena commonly encountered in computational mechanics. Across all benchmarks, SEDONet consistently achieves the lowest or statistically comparable relative $L^2$ errors among DeepONet, FEDONet, and SEDONet, with improvements of up to 54\% over the baseline DeepONet and meaningful gains over Fourier-embedded variants for bounded, non-periodic problems. Energy spectrum analyses further demonstrate that SEDONet more accurately preserves intermediate- and high-frequency solution structures. In addition, the proposed Chebyshev embedding is successfully extended to the Fourier Neural Operator (FNO), indicating that the underlying spectral representation is not restricted to the DeepONet architecture. The proposed framework provides a simple, parameter-neutral modification to DeepONets, delivering a robust and computationally efficient spectral framework for surrogate modeling of nonlinear operators arising in scientific computing.

\end{abstract}


\section{Introduction}\label{sec:intro}

Partial differential equations (PDEs) serve as the foundation for modelling a wide range of physical, biological and engineering systems, including diffusion, transport, turbulence, phase transitions and chaotic dynamics.  
Classical numerical methods, finite-difference, finite-element, and spectral discretizations, provide high-fidelity solutions but are computationally expensive when repeatedly solving parametric or high-dimensional PDEs, or when rapid surrogate evaluations are required \cite{boyd2001chebyshev,Gottlieb1977spectral,Press1986numericalrecipes}.  
To accelerate PDE modelling pipelines, a long line of research has investigated surrogate approximations, including universal neural approximators \cite{cybenko1989approximation,hornik1989multilayer,58326,chen1995universal}, reduced-order models such as POD \cite{sirovich1987turbulence,berkooz1993proper}, Gaussian process regression \cite{10.7551/mitpress/3206.001.0001}, and mesh-free radial-basis approaches \cite{KANSA1990127,lowe1988multivariable}.  
While successful on fixed grids, these classical surrogates do not naturally extend to operator learning, the task of learning mappings between infinite-dimensional function spaces.

\medskip
Neural operators address this challenge by learning solution operators directly from data.  
Deep Operator Networks (DeepONets) \cite{lu2021deeponet} and Neural Operators \cite{10.5555/3648699.3648788} define branch-trunk factorizations that generalise across discretisations and provide operator-level universal approximation guarantees.  
Fourier Neural Operators (FNOs) extend this idea using global spectral convolutions and have demonstrated strong performance on fluid and parametric PDEs \cite{li2021fourierneuraloperatorparametric}.  
Subsequent variants integrate multi-wavelets \cite{TRIPURA2023115783,gupta2021multiwaveletbased}, hierarchical tensorisation \cite{kossaifi2023multi,guo2024mgfnomultigridarchitecturefourier}, graph kernels \cite{li2020neuraloperatorgraphkernel,li2020multipolegraphneuraloperator}, and geometry-aware deformations \cite{li2023fourier,li2023geometry,huang2025operator}, further enriching the operator-learning landscape.  
Recent work has also emphasised closure modelling and multifidelity formulations in multiscale settings, notably the multifidelity DeepONet approaches of Ahmed and Stinis and co-workers \cite{ahmed2023multifidelitydeeponet,howard2023multifidelitydeeponet}.  
Kernel and Gaussian-process baselines have been revisited in this context, showing that carefully designed kernel methods can be competitive with neural operators \cite{batlle2024kerneloperatorlearning} and that neural-operator-induced Gaussian processes provide a probabilistic, uncertainty-aware extension of deterministic operators \cite{kumar2024nogap}.

\medskip
Beyond these foundational architectures, operator learning has recently expanded toward more specialized and physically informed formulations. New directions include neuroscience-inspired neural operators that strengthen multiscale feature representation~\cite{garg2024neuroscienceNO}, derivative-informed architectures that embed local sensitivity structure~\cite{oleary2024dino}, and methods designed to mitigate spectral bias in stiff or multiscale PDE regimes~\cite{liu2024spectralbias}. Other developments target challenging physical settings such as interface-dominated problems, where interface-aware operator networks improve discontinuity resolution~\cite{wu2024ionet,bi2025xideeponet}, and scenarios involving geometric or resolution variability, where local and resolution-invariant operator formulations enhance generalization across meshes and spatial discretizations~\cite{li2024localNO,huang2025resolutionDON}. Complementary research leverages dynamical-systems structure through Koopman-based operator-learning models~\cite{meng2024koopmanINN}, while nonlocal material behaviour has motivated peridynamic operator networks~\cite{jafarzadeh2024peridynamicNO}. Additional work explores feature-adjacent mappings to stabilize learning in heterogeneous environments~\cite{chen2024featureadjacentMF}. Together these advances illustrate a clear shift toward domain-informed, physics-aware neural operators that better capture complex PDE phenomena and broaden the applicability of operator-learning frameworks.

\medskip
A rich body of work has further explored specialized neural operators for complex physical phenomena and non-standard constitutive behaviour.  
Peridynamic neural operators provide nonlocal constitutive models for material deformation \cite{jafarzadeh2024peridynamicneuraloperators}, while probabilistic closure models combining conditional diffusion processes with neural operators have been proposed for stochastic turbulence modeling \cite{dong2025stochasticclosure}.  
Resolution-independent neural operators \cite{bahmani2025rino} and local neural operators \cite{li2024localneuraloperator} address mesh and domain-variability, enabling robust generalization across resolutions and geometries.  
Physics-informed geometry-aware neural operators \cite{zhong2023pigano} and Cole--Hopf-based operator constructions \cite{xu2025colehopfoperator} continue to illustrate how classical analytical transformations can enhance neural operator architectures.

\medskip
In parallel, there has been rapid progress in physics-informed and probabilistic operator learning.  
Physics-informed DeepONets and neural operators \cite{goswami2022physicsinformeddeepneuraloperator,li2023physicsinformedneuraloperatorlearning,eshaghi2024variationalphysicsinformedneuraloperator,chen2025pseudo} incorporate PDE residuals and variational principles into the training objective, while latent neural operators and invertible Fourier neural operators introduce expressive latent spaces and reversible dynamics \cite{wang2024latentneuraloperatorsolving,wang2024latentneuraloperatorpretraining,ahmad2024diffeomorphiclatentneuraloperators,long2025invertiblefourierneuraloperators}.  
Sobolev-style training strategies \cite{cho2025sobolevoperator} and unsupervised operator learning for mean-field games \cite{huang2025unsupervisedmfg} highlight the role of derivative information and weak formulations in improving generalization.  
At the same time, in-context generalization properties of operator networks are being explored through PDE-oriented in-context models \cite{yang2024iconpde}.  
Collectively, these developments indicate a maturing ecosystem where neural operators, Gaussian processes, and hybrid architectures are increasingly unified.

\medskip
Despite these advances, the trunk network in DeepONet, responsible for encoding spatial or spatiotemporal coordinates, is typically a shallow multilayer perceptron (MLP).  
MLPs exhibit a well-known spectral bias, that preferentially learn low-frequency components while struggling with high-frequency, oscillatory or multiscale patterns \cite{rahman2019spectralbiasneuralnetworks}.  
This significantly affects PDEs with boundary layers, steep gradients, or sharp phase-field transitions.  
In Euclidean neural networks, this issue is often mitigated by positional encodings such as sinusoidal/Fourier features \cite{tancik2020fourierfeaturesletnetworks} or learned embeddings \cite{kast2024positional}.  
In particular, Fourier-Embedded DeepONet (FEDONet) \cite{sojitra2025fedonet} incorporates Fourier features into DeepONet trunks, improving performance on periodic or nearly periodic PDEs.  
However, Fourier features encode a fundamentally periodic prior and are therefore less suited to bounded, non-periodic domains common in elliptic and parabolic PDEs, motivating alternative non-periodic or polynomial spectral priors such as those explored in \cite{liu2024orthogonalpolynomialneuraloperator,bahmani2025rino,li2024localneuraloperator}.

\medskip
Chebyshev polynomials provide an orthogonal basis on bounded intervals and form the cornerstone of polynomial spectral methods \cite{boyd2001chebyshev,Gottlieb1977spectral}.  
They offer exponential convergence for smooth solutions, excellent resolution of boundary layers, and natural compatibility with non-periodic boundary conditions.  
A growing literature explores Chebyshev-based neural representations, including Chebyshev neural networks \cite{mall2017singlechebyshev}, fractional PDE solvers \cite{sivalingam2024chebyshevNNfractional}, Chebyshev spectral neural networks \cite{yin2024chebyshevspectralNN}, orthogonal polynomial neural operators \cite{liu2024orthogonalpolynomialneuraloperator}, Chebyshev-enhanced PINNs \cite{huang2025cdpinn,chen2025chebyshevsobolevpinn}, hybrid Chebyshev–attention architectures \cite{zhang2024acpkan}, and Chebyshev feature networks \cite{xu2025chebyshevfeatureNN}.  
Spectral neural operators combining Fourier and Chebyshev representations have also been proposed \cite{fanaskov2022spectral}.  
These studies suggest that embedding Chebyshev structure inside modern neural PDE solvers can substantially improve stability and accuracy on non-periodic domains.

\medskip
However, existing Chebyshev-based neural architectures either parametrize the solution directly in spectral space, or operate on coefficient maps rather than spatial coordinates.  
None provide a simple, drop-in non-periodic spectral embedding for DeepONet trunks analogous to Fourier embeddings in FEDONet.  
To bridge this gap, we propose the Spectral-Embedded DeepONet (SEDONet), a Chebyshev-based operator network in which raw coordinates are replaced by a fixed tensor-product Chebyshev embedding.  
The embedding supplies a structured polynomial basis aligned with bounded, non-periodic domains.  
The branch network remains unchanged, while the trunk network mixes Chebyshev modes into adaptive basis functions.  
SEDONet introduces no additional trainable parameters, preserves the DeepONet factorization, and is compatible with existing DeepONet/FEDONet implementations.

\medskip
We evaluate SEDONet on the following six benchmark systems:
\begin{itemize}
    \item 2-D Poisson equation (elliptic),
    \item 1-D viscous Burgers' equation (nonlinear transport-diffusion),
    \item 1-D advection-diffusion equation (Dirichlet boundary conditions),
    \item Lorenz--96 chaotic ODE system,
    \item Allen--Cahn phase-field equation (interface dynamics), and
    \item 2-D Darcy flow (steady-state elliptic flow in heterogeneous porous media).
\end{itemize}

These benchmarks span elliptic, hyperbolic, parabolic, phase-field, and chaotic systems, providing a broad testbed that encompasses boundary layers, shocks, interface dynamics, and multiscale solution behavior. Across all benchmark problems, SEDONet consistently achieves lower relative $L^2$ errors than both DeepONet and FEDONet, with the most significant improvements observed for bounded, non-periodic PDEs. The primary objective of this study is to investigate Chebyshev-based spectral embeddings within the DeepONet framework and systematically compare their performance with conventional Fourier-based embeddings.

Our work makes three primary contributions, which are summarized as follows. First, we introduce a Chebyshev-embedded trunk for DeepONet, providing a plug-and-play spectral coordinate embedding tailored to non-periodic PDE domains. This design preserves the original branch-trunk factorization while equipping the trunk with a principled polynomial spectral prior. Second, we provide a theoretical analysis showing that the resulting Chebyshev features form an approximately orthogonal and well-conditioned Gram matrix, thereby expanding the hypothesis class of coordinate-input DeepONets and improving their ability to represent functions with boundary layers or steep gradients. Third, through extensive experiments on six diverse benchmark problems spanning elliptic, hyperbolic, parabolic, nonlinear transport, chaotic ODE dynamics, phase-field models, and Darcy flow, we demonstrate that SEDONet consistently improves reconstruction accuracy while maintaining discretization invariance and low model complexity. Furthermore, we show that the proposed Chebyshev embedding can be successfully extended beyond the DeepONet architecture to the Fourier Neural Operator (FNO), highlighting the generality of the proposed spectral representation. Together, these developments embed the structure of classical non-periodic spectral methods directly into the trunk of DeepONet, extending operator learning beyond Fourier-based embeddings and offering a natural bridge between polynomial spectral approximation theory and modern neural operator architectures.

\medskip
The remainder of this paper is organized as follows. Section~\ref{Methodology} presents the formulation of the Spectral-Embedded DeepONet (SEDONet), detailing the Chebyshev spectral embedding, its integration into the trunk network, and the resulting spectral interpretation for operator approximation on bounded domains. 
Section~\ref{Results} describes the benchmark setup, dataset generation, and implementation details across the six representative operator families considered, including elliptic, parabolic, advective-diffusive, chaotic, and phase-field dynamics. 
Section~\ref{Summary} reports the quantitative and qualitative results, highlighting the improvements achieved by SEDONet over DeepONet and FEDONet in reconstruction accuracy, boundary resolution, and spectral fidelity. 
Finally, Section~\ref{Futurework} concludes the study and outlines several promising research directions in spectral operator learning, adaptive polynomial embeddings, and physics-informed neural operator architectures.

\section{Methodology}
\label{Methodology}

\subsection{Problem Formulation}
\label{subsec:problem_formulation}
We consider the problem of learning nonlinear operators between
infinite-dimensional function spaces. Let $\Omega \subset \mathbb{R}^D$
be a bounded domain, and define the input and output function spaces
\begin{align}
    \mathcal{U} &= \{ u : \mathcal{X} \to \mathbb{R}^{d_u} \}, 
    \qquad \mathcal{X} \subseteq \mathbb{R}^{d_x},\\
    \mathcal{S} &= \{ s : \mathcal{Y} \to \mathbb{R}^{d_s} \}, 
    \qquad \mathcal{Y} \subseteq \mathbb{R}^{d_y}.
\end{align}
We assume access to paired samples
$\mathcal{D}=\{(u^i,s^i)\}_{i=1}^N$ from an unknown operator
$\mathcal{G}:\mathcal{U}\to\mathcal{S}$, e.g., the solution operator of
a PDE mapping initial or forcing data to a full space-time field.  
Our goal is to learn a parametric approximation
$\mathcal{G}_\theta:\mathcal{U}\to\mathcal{S}$ that generalizes to new,
unseen functions $u\in\mathcal{U}$.

Deep Operator Networks (DeepONets) provide a natural architecture for
this task. Each prediction decomposes into a branch-trunk factorization
of the form
\begin{equation}
    \mathcal{G}_\theta(u)(\zeta)
    =
    B_\theta(u)\,\cdot\,T_\theta(\zeta), 
    \qquad \zeta=(x,y,z,t),
\end{equation}
where the branch network $B_\theta$ encodes discrete samples of the
input function $u$ and the trunk network $T_\theta$ encodes coordinate
queries $\zeta$. This separation mirrors classical spectral
representations: the branch returns coefficients, while the trunk
returns basis functions evaluated at $\zeta$.
Consequently, the choice of trunk representation is crucial for the
model’s spectral expressivity and its ability to represent different
PDE structures.

\subsection{Spectral Embeddings for Non-Periodic Operators}
\label{subsec:spectral_embeddings}

In the standard DeepONet formulation, $T_\theta(\zeta)$ is produced by a
multi-layer perceptron (MLP) acting directly on the raw coordinates.
This simple parameterization is universal in principle, but in practice
it exhibits a strong spectral bias toward low frequencies and often
struggles with non-periodic geometries, sharp gradients, and boundary
layers. Coordinate encodings based on sinusoidal features (e.g., Fourier
features or positional embeddings) can partially alleviate this issue,
but they are most naturally tailored to periodic domains.

Many operators arising in scientific computing are instead defined on
bounded intervals with Dirichlet or Neumann boundary conditions.  
Examples include elliptic Poisson problems, reaction-diffusion systems,
advective-diffusive flows, and chaotic ODEs (when interpreted as
operators on a finite time window). For such settings, polynomial
spectral methods, in particular Chebyshev expansions, provide
well-conditioned, rapidly convergent bases on $[-1,1]$ that naturally
resolve boundary layers and steep gradients.

These observations motivate a trunk architecture that embeds coordinates
into a non-periodic spectral dictionary before applying a
learned MLP. The resulting model, which we call the
Spectral-Embedded DeepONet (SEDONet), combines the flexibility of
DeepONets with the approximation properties of Chebyshev spectral
methods.
\subsection{Relation to Existing Chebyshev-Based Neural Networks}
\label{subsec:existing_chebyshev}
Chebyshev polynomials have recently been incorporated into a variety of neural network architectures for scientific machine learning, including Chebyshev neural networks for function approximation \cite{mall2017singlechebyshev,xu2025chebyshevfeatureNN}, Chebyshev spectral neural networks for solving partial differential equations \cite{yin2024chebyshevspectralNN}, Chebyshev-based physics-informed neural networks \cite{huang2025cdpinn,chen2025chebyshevsobolevpinn}, and orthogonal polynomial neural operators for non-periodic PDEs \cite{liu2024orthogonalpolynomialneuraloperator}. In these approaches, Chebyshev polynomials are primarily employed to improve the approximation of finite-dimensional functions, construct spectral representations of the solution, or parameterize polynomial coefficient mappings. Consequently, the underlying neural network learns a mapping between finite-dimensional input and output variables.

The role of the trunk network in DeepONet is fundamentally different from that of a conventional multilayer perceptron (MLP). Rather than approximating a finite-dimensional function, DeepONet learns nonlinear operators that map between infinite-dimensional function spaces \cite{lu2021deeponet,10.5555/3648699.3648788}. Within the branch-trunk decomposition, the branch network encodes the input function into latent coefficients, whereas the trunk network evaluates coordinate-dependent basis functions at arbitrary spatial or spatio-temporal locations. The operator prediction is then obtained through the interaction of these two components, enabling DeepONet to approximate solution operators instead of individual functions.

From this perspective, introducing Chebyshev embeddings into the trunk network does not simply replace the input features of a conventional MLP. Instead, it modifies the coordinate representation used to construct the operator basis itself. The deterministic Chebyshev dictionary provides a structured polynomial basis tailored to bounded, non-periodic domains before the trunk network learns adaptive basis functions. Consequently, the proposed embedding changes the functional representation employed by the operator-learning framework while preserving the original branch-trunk factorization of DeepONet.

Therefore, the proposed Spectral-Embedded DeepONet (SEDONet) should be viewed as an operator-level spectral embedding strategy rather than a conventional Chebyshev neural network. Unlike previous Chebyshev-based architectures, which either enhance conventional neural networks for function approximation or develop operator-learning frameworks based on polynomial representations \cite{mall2017singlechebyshev,yin2024chebyshevspectralNN,xu2025chebyshevfeatureNN,liu2024orthogonalpolynomialneuraloperator}, SEDONet specifically introduces Chebyshev polynomial embeddings into the trunk network of DeepONet. This preserves the original branch-trunk factorization while modifying only the coordinate representation used to construct the operator basis. Furthermore, unlike FEDONet \cite{sojitra2025fedonet}, which introduces Fourier feature embeddings to encode periodic spectral information, SEDONet replaces this periodic prior with a Chebyshev polynomial dictionary that is naturally aligned with bounded, non-periodic domains. Consequently, SEDONet provides a more suitable inductive bias for operator learning on bounded domains while remaining fully compatible with the original DeepONet architecture.

\subsection{Model Complexity}
\label{subsec:model_complexity}
An important consideration when introducing spectral embeddings is whether the improved predictive performance is achieved at the expense of increased model complexity. Unlike modifying the network architecture by introducing additional trainable layers or substantially increasing the network width, the proposed Chebyshev embedding is a deterministic feature transformation applied to the input coordinates before they are processed by the trunk network. Consequently, the branch-trunk architecture of DeepONet remains unchanged, and the embedding itself does not introduce additional trainable parameters.

To ensure a fair comparison, all experiments presented in this work employ identical branch and trunk network configurations for DeepONet, FEDONet, and SEDONet, with matched network depths and widths. Therefore, the overall number of trainable parameters remains comparable across all three models. As a result, the observed improvements in predictive accuracy cannot be attributed to increased model capacity but rather to the enhanced coordinate representation provided by the proposed Chebyshev polynomial embedding.

Furthermore, similar to the Fourier embedding adopted in FEDONet, the proposed Chebyshev embedding acts as a fixed spectral preconditioning of the input coordinates while preserving the original DeepONet formulation. Consequently, SEDONet maintains essentially the same architectural complexity as both the baseline DeepONet and FEDONet, providing a simple and effective modification that is readily compatible with existing DeepONet implementations.

\subsection{Spectral-Embedded DeepONet (SEDONet): 
Chebyshev Trunk for Bounded Domains}
\label{subsec:sedonet}

SEDONet retains the classical DeepONet branch but augments the trunk
with a deterministic Chebyshev polynomial dictionary. Instead of feeding
raw coordinates to the trunk MLP, we first compute fixed spectral
features
\begin{equation}
    \phi_{\text{Cheb}}(y)\in\mathbb{R}^{d_{\text{trunk}}},
    \qquad y=(x,t),
\end{equation}
and then learn a nonlinear mapping from these features to trunk outputs.
The resulting model can be interpreted as a data-driven Chebyshev
expansion adapted to bounded, non-periodic PDE domains.

In one-dimensional Chebyshev basis, for a spatial coordinate $x\in[0,1]$, we apply the affine transformation
\begin{equation}
    \xi = 2x - 1 \in [-1,1],
\end{equation}
and work with the standard Chebyshev polynomials of the first kind
$\{T_n\}_{n\ge 0}$ defined by the recurrence
\begin{equation}
    T_0(\xi) = 1,\qquad
    T_1(\xi) = \xi,\qquad
    T_{n+1}(\xi) = 2\xi\,T_n(\xi) - T_{n-1}(\xi).
    \label{eq:cheb_recursion}
\end{equation}
These polynomials form an orthogonal basis on $[-1,1]$ with respect to
the weight $w(\xi) = (1-\xi^2)^{-1/2}$
(see Appendix~\ref{appendix:cheb-spectral} for details).  
Orthogonality and the clustering of Chebyshev nodes near the endpoints
make this basis particularly effective for representing non-periodic PDE
solutions with boundary layers and steep gradients.


In the tensor-product spectral dictionary, for spatio-temporal inputs $y = (x,t)\in [0,1]^2$, we map
\begin{equation}
    \xi_x = 2x - 1, \qquad \xi_t = 2t - 1,
\end{equation}
and construct a tensor-product dictionary a follows
\begin{equation}
    \Phi_{ij}(x,t)
    = T_i(\xi_x)\, T_j(\xi_t), \qquad
    0\le i < K_x,\quad 0\le j < K_t.
    \label{eq:cheb_tensor}
\end{equation}
Collecting all entries in lexicographic order yields the full
deterministic feature vector of the following form as
\begin{equation}
    \Phi_{\text{full}}(x,t)
    =
    \bigl[\Phi_{00},\Phi_{01},\dots,\Phi_{K_x-1,K_t-1}\bigr]^\top
    \in \mathbb{R}^{K_x K_t}.
    \label{eq:cheb_full_vec}
\end{equation}
This tensor-product construction extends immediately to higher spatial
dimensions by adding additional Chebyshev factors.

For the fixed-width Chebyshev embedding, the common practice, the trunk network expects a fixed input dimension
$d_{\text{trunk}}$ that may differ from $K_x K_t$.  
To bridge this gap we introduce a simple crop/pad operator
$C:\mathbb{R}^{K_xK_t}\to\mathbb{R}^{d_{\text{trunk}}}$ and define the
final embedding as
\begin{equation}
    \phi_{\text{Cheb}}(x,t)
    =
    C\,\Phi_{\text{full}}(x,t).
    \label{eq:cheb_embedding}
\end{equation}
The operator $C$ can either truncate higher-order modes or pad with
zeros, depending on the chosen spectral resolution.  
Importantly, $\phi_{\text{Cheb}}$ is deterministic and fixed,
no trainable parameters are attached to the embedding itself, so the
inductive spectral structure remains stable during optimization.

For the trunk network, the SEDONet trunk maps Chebyshev features to $p$ latent basis channels as the followings
\begin{equation}
    T_\theta(x,t)
    =
    \Psi_\theta\!\left(\phi_{\text{Cheb}}(x,t)\right)
    =
    [t_1(x,t),\dots,t_p(x,t)]^\top
    \in\mathbb{R}^{p},
    \label{eq:sedonet_trunk}
\end{equation}
where $\Psi_\theta$ is an MLP.
The functions $t_k(x,t)$ can be viewed as learned basis functions
formed by nonlinear modulation of Chebyshev modes.

In the branch network and operator synthesis, the branch network mirrors the standard DeepONet branch.  
Given discrete samples of an input function $u_0$, it produces latent
coefficients
\begin{equation}
    B_\theta(u_0)
    =
    [b_1(u_0),\dots,b_p(u_0)]^\top .
    \label{eq:sedonet_branch}
\end{equation}
The predicted solution at any coordinate $(x,t)$ is then synthesized via
the DeepONet rule as followings
\begin{equation}
    \widehat{u}(x,t;u_0)
    =
    B_\theta(u_0)^\top T_\theta(x,t)
    =
    \sum_{k=1}^{p} b_k(u_0)\, t_k(x,t).
    \label{eq:sedonet_pred}
\end{equation}
This expression highlights the analogy with classical spectral methods, the branch computes generalized ``spectral coefficients'' from $u_0$,
while the trunk provides basis functions evaluated at $(x,t)$. For the training objective, given training pairs $(u_0^{(i)},u^{(i)})$ evaluated at points
$\{(x^q,t^q)\}_{q=1}^Q$, we train SEDONet by minimizing the empirical
mean-squared error as follows
\begin{equation}
    \mathcal{L}_{\text{SEDONet}}(\theta)
    =
    \frac{1}{NQ}
    \sum_{i=1}^{N}
    \sum_{q=1}^{Q}
    \bigl|
        B_\theta(u_0^{(i)})^\top T_\theta(x^q,t^q)
        - u^{(i)}(x^q,t^q)
    \bigr|^2 .
    \label{eq:sedonet_loss}
\end{equation}
We employ mini-batch Adam optimization.  
All trainable parameters reside in the branch and trunk MLPs; the
Chebyshev embedding is fixed and non-learnable, acting purely as an
inductive bias. For the spectral interpretation, it is convenient to view SEDONet in a spectral framework.  
Suppose the trunk and branch can be written as
\[
    T_\theta(x,t)=[\psi_1(x,t),\dots,\psi_p(x,t)]^\top,\qquad
    B_\theta(u_0)=[\langle u_0,\varphi_1\rangle,\dots,\langle u_0,\varphi_p\rangle]^\top,
\]
for some learned feature functionals $\varphi_k$ and also the basis functions
$\psi_k$.  
Then the prediction admits the form as
\begin{equation}
    \widehat{\mathcal{G}}(u_0)(x,t)
    \approx
    \sum_{k=1}^{p}
    \langle u_0,\varphi_k\rangle\,\psi_k(x,t),
    \label{eq:sedonet_spectral_decomp}
\end{equation}
i.e., a data-driven Chebyshev-type spectral expansion.  
Compared to vanilla coordinate-input trunks, the hypothesis space
induced by Chebyshev embeddings strictly enlarges the class of
representable functions on bounded domains
(see Appendix~\ref{appendix:cheb-superset}).

\subsection{Training and Evaluation Protocol}
\label{subsec:sedonet_eval}

After defining the architecture, we train SEDONet, DeepONet, and FEDONet under a common protocol and compare their performance on six benchmark families, including the 2-D Poisson equation, 1-D Burgers' equation, 1-D advection-diffusion equation, Allen-Cahn equation, Lorenz-96 chaotic system, and 2-D Darcy flow. During training, each mini-batch consists of a set of input functions
$\{u_0^{(i)}\}$ and corresponding reference solutions
$\{u^{(i)}(x^q,t^q)\}$ sampled on dense grids.  
For each architecture, we compute the branch coefficients, trunk
outputs, and synthesized predictions via
\eqref{eq:sedonet_pred}, and update parameters by minimizing the loss
\eqref{eq:sedonet_loss} (or its DeepONet/FEDONet counterparts). On held-out test data, we use several complementary diagnostics, for a given test input $u$, we evaluate the relative $L^2$ error as
\begin{equation}
    \varepsilon_{L^2}(u)
        =
        \frac{\|\mathcal{G}_\theta(u)-\mathcal{G}(u)\|_2}
             {\|\mathcal{G}(u)\|_2},
\end{equation}
computed on a dense evaluation grid tailored to each benchmark.  
This metric quantifies the normalized discrepancy between the reference
and predicted fields, and forms the basis of the summary statistics
reported in Table~\ref{tab:relL2} and the error-bar plots. For the spectral fidelity for PDE benchmarks, we additionally compute angle-integrated power
spectra of the reference and predicted solutions and compare
$E_{\text{pred}}(k)$ to $E_{\text{ref}}(k)$.  
This reveals how well each architecture captures multiscale structure
and high-frequency content, beyond what is visible in a single scalar
error.

Finally, for the qualitative diagnostics, we visualize representative solution fields and corresponding
error maps.  
These plots highlight model behavior near boundaries, interfaces, and
nonlinear structures where baseline methods are known to deteriorate.
The qualitative observations are reported in Section~\ref{Results}
and closely track the quantitative trends. This unified protocol enables a fair comparison of spectral resolution,
boundary behavior, and generalization performance across all three
operator-learning architectures.

\subsection{Algorithmic Workflow}
\label{subsec:workflow}
\begin{figure}[ht!]
    \centering
    \includegraphics[width=.9\linewidth]{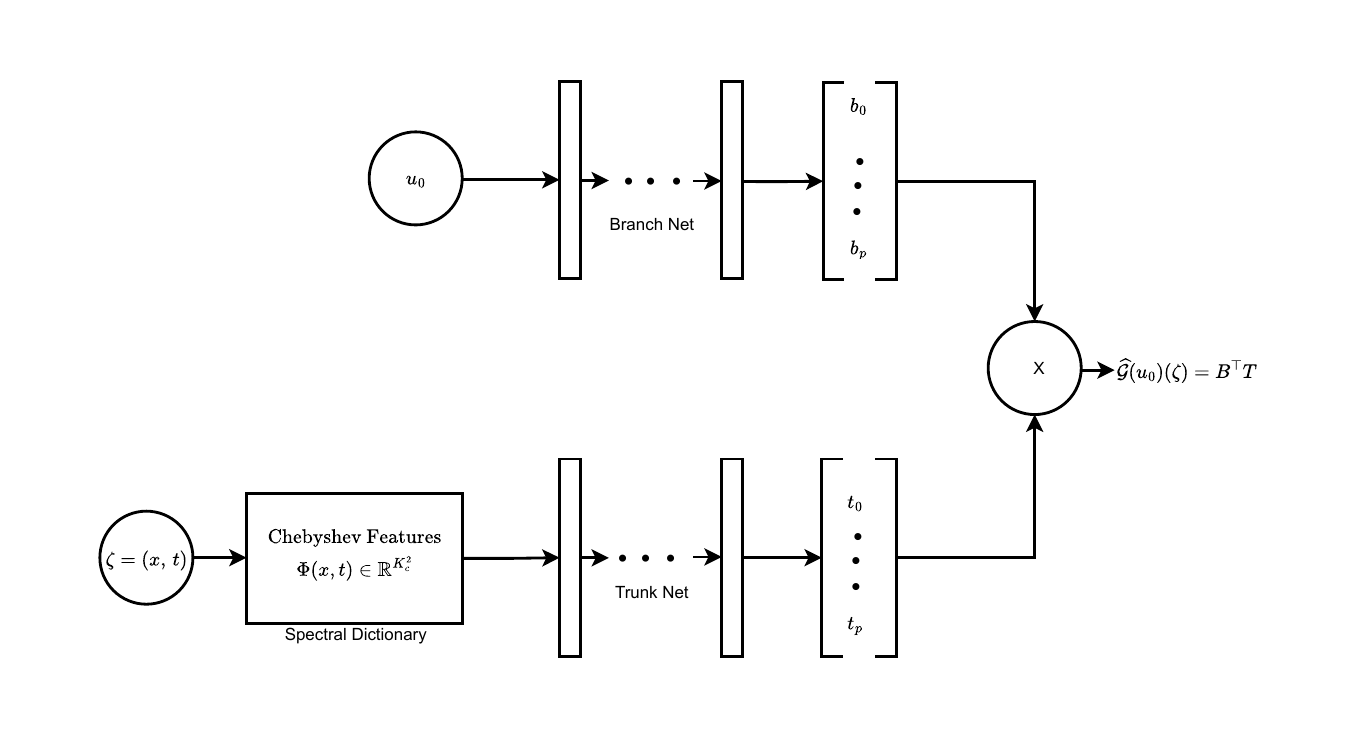}
    \caption{\textbf{SEDONet architecture:}
The branch network maps discrete samples of the input function $u_0$ to latent coefficients $b_k$, while the trunk network maps Chebyshev spectral features $\Phi(x,t)$ to basis channels $t_k(x,t)$. Their inner product yields the operator evaluation $\widehat{\mathcal{G}}(u_0)(x,t)$.}
    \label{fig:sedonet_cheb}
\end{figure}

The learning procedure for SEDONet follows the general DeepONet framework but replaces raw coordinate inputs with the Chebyshev spectral embedding introduced in Section~\ref{subsec:sedonet}.  
This results in a branch-trunk architecture in which the branch network encodes the input function $u_0$, and the trunk network evaluates Chebyshev-modulated basis functions at any spatio-temporal coordinate $(x,t)$.  
Figure~\ref{fig:sedonet_cheb} provides a schematic overview of this structure, illustrating how the branch and trunk outputs are combined through an inner product to produce operator evaluations.

\begin{algorithm}[ht!]
\caption{Training procedure for SEDONet}
\label{alg:sedonet}
\begin{algorithmic}[1]
\Require Training set $\mathcal{D} = \{(u_0^{(i)}, u^{(i)})\}_{i=1}^N$,
         coordinate grid $\{(x^q,t^q)\}_{q=1}^Q$,
         number of trunk channels $p$,
         learning rate $\eta$,
         number of epochs $E$
\Ensure Trained parameters $\theta = (\theta_{\text{branch}},\theta_{\text{trunk}})$
\State Initialize $\theta_{\text{branch}}, \theta_{\text{trunk}}$
\For{$e = 1,\dots,E$}
    \For{each mini-batch $\mathcal{B} \subset \mathcal{D}$}
        \State Sample input functions $\{u_0^{(i)}\}_{i\in\mathcal{B}}$
               and reference values
               $\{u^{(i)}(x^q,t^q)\}_{i\in\mathcal{B}, q=1,\dots,Q}$
        \State \textbf{Branch step:}
               compute latent coefficients
               $B_\theta(u_0^{(i)}) \in \mathbb{R}^p$
               for all $i$
        \State \textbf{Trunk step:}
               for each grid point $(x^q,t^q)$,
               evaluate Chebyshev embedding
               $\phi_{\text{Cheb}}(x^q,t^q)$
               and trunk output
               $T_\theta(x^q,t^q)\in\mathbb{R}^p$
        \State \textbf{Synthesis:}
               form predictions
               $\widehat{u}^{(i)}(x^q,t^q)
                 = B_\theta(u_0^{(i)})^\top T_\theta(x^q,t^q)$
        \State \textbf{Loss:}
               compute $\mathcal{L}_{\text{SEDONet}}(\theta)$
               using \eqref{eq:sedonet_loss}
        \State \textbf{Update:}
               apply one Adam step with learning rate $\eta$
               to update $\theta_{\text{branch}},\theta_{\text{trunk}}$
    \EndFor
\EndFor
\end{algorithmic}
\end{algorithm}
During training, SEDONet alternates between three stages for each mini-batch. Firstly, encoding the input function through the branch net to obtain latent coefficients,  then secondly evaluating the Chebyshev embedding and trunk net at all query points, and  lastly synthesizing predictions by contracting the branch and trunk outputs, followed by gradient-based parameter updates.  
Because the Chebyshev embedding is fixed, the trunk network receives a stable spectral dictionary, which improves conditioning and accelerates convergence compared to purely coordinate-based trunks.  
Algorithm~\ref{alg:sedonet} provides a detailed summary of this workflow.

\medskip
As illustrated in Figure~\ref{fig:sedonet_cheb}, the division of labour between the branch and trunk networks results in a flexible yet structured operator-learning pipeline. The branch network captures the dependence of the solution on the input function, while the Chebyshev-enhanced trunk network provides a spectrally rich representation of the output domain. This separation enables SEDONet to generalize across discretizations and spatial resolutions while retaining the favorable approximation properties of Chebyshev spectral methods on bounded, non-periodic domains.


\section{Results}
\label{Results}
In this section, we evaluate the proposed SEDONet across a diverse set of benchmark problems, including the two-dimensional Poisson equation, Burgers' equation, the Lorenz-96 system, the Allen-Cahn equation, the advection-diffusion equation, and Darcy flow in a rectangular domain. For each benchmark, we compare the predictive performance of the baseline DeepONet, FEDONet, and the proposed SEDONet using multiple complementary evaluation criteria. Specifically, we report the mean relative $L^2$ error over the test set, present qualitative comparisons of the reconstructed solutions on representative test examples, and analyze the corresponding error distributions. In addition, to assess how accurately each model captures the underlying spectral characteristics of the solution, we examine the energy spectra of the predicted fields. While the relative $L^2$ error provides a measure of the overall reconstruction accuracy, the energy spectrum offers additional insight into the representation of spatial frequencies, particularly the intermediate- and high-frequency components that govern fine-scale solution structures. Together, these quantitative and qualitative evaluations provide a comprehensive assessment of the proposed method across a range of PDEs.

\subsection{2D Poisson Equation}
\label{sec:poisson}

We evaluate all three models, DeepONet, FEDONet, and the proposed SEDONet, on the two-dimensional Poisson equation
\begin{equation}
\nabla^2 u(x,y) = f(x,y), \qquad (x,y) \in [0,1]^2,
\end{equation}
subject to homogeneous Dirichlet boundary conditions,
\begin{equation}
u(x,y)\big|_{\partial\Omega} = 0.
\end{equation}
The forcing field $f(x,y)$ serves as input, and $u(x,y)$ denotes the steady-state solution. A dataset of $N=10{,}000$ operator pairs is constructed by sampling forcing fields from a Gaussian Random Field,
\[
f \sim \mathrm{GRF}(\alpha = 3,\, \tau = 3),
\]
on a uniform $128\times128$ grid and solving the Poisson problem using a five-point finite-difference discretization. This yields the linear system
\[
A\mathbf{u} = \mathbf{b},
\]
where $\mathbf{b}$ is the vectorized forcing and $\mathbf{u}$ is the numerical solution. The matrix $A$ follows the standard discrete Laplacian stencil,
\[
A_{ij} =
\begin{cases}
4,  & i=j, \\
-1, & \text{if nodes } i \text{ and } j \text{ share an edge}, \\
0,  & \text{otherwise},
\end{cases}
\]
with boundary rows modified to impose the homogeneous Dirichlet conditions.

\begin{figure}[H]
    \centering
    \includegraphics[width=0.93\linewidth]{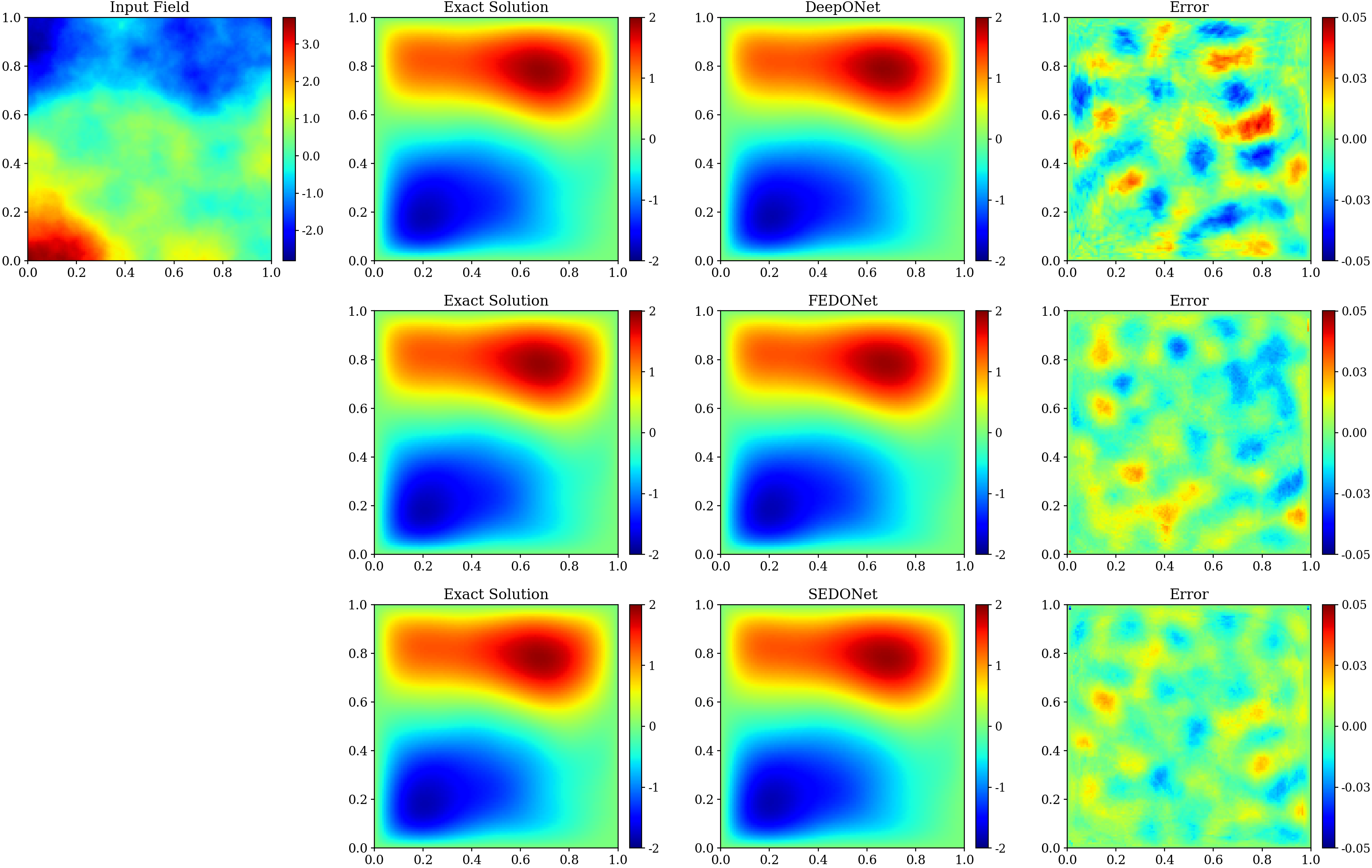}
    \caption{Comparison of DeepONet, FEDONet, and SEDONet on a representative test example from the 2D Poisson dataset.
    The left panel shows the forcing field; the remaining panels show the exact solution, model predictions, and the corresponding pointwise residual fields.}
    \label{fig:2dpoisson_field_comp}
\end{figure}

Figure~\ref{fig:2dpoisson_field_comp} presents a representative test example. The left panel displays the input forcing field, followed by the exact solution and the predictions from DeepONet, FEDONet, and SEDONet, together with their pointwise residual fields. DeepONet captures the overall structure but exhibits smoothing of interior features and larger residual magnitudes. FEDONet improves spatial fidelity through Fourier embeddings. SEDONet achieves the most accurate reconstruction: its Chebyshev-based spectral embedding aligns well with bounded domains, enabling it to resolve both interior variations and boundary gradients more effectively. Furthermore, the residual field remains smoothly distributed throughout the computational domain without noticeable error concentration near the boundaries, indicating that the proposed embedding preserves both the global solution structure and the boundary behavior.

\begin{figure}[h!]
    \centering
    \includegraphics[width=0.60\linewidth]{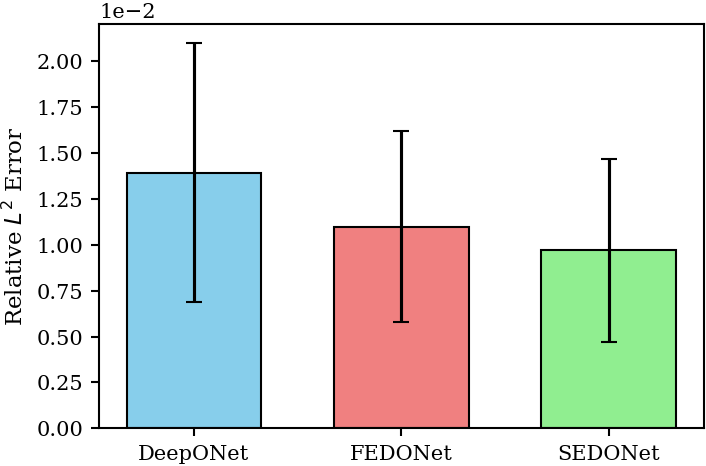}
    \caption{Relative $L^2$ error (mean $\pm$ std) across 1000 unseen Poisson test sets for all three architectures.
    SEDONet achieves both the lowest mean error and the lowest variance.}
    \label{fig:poisson2d_relative_l2}
\end{figure}

These qualitative observations are consistent with the quantitative results. Across the full test set, DeepONet attains a mean relative $L^2$ error of $1.39\%$, FEDONet reduces this to $1.10\%$, and SEDONet further improves the accuracy to $\mathbf{0.97\%}$. This corresponds to improvements of $31.2\%$ over DeepONet and $11.0\%$ over FEDONet. The bar plot in Figure~\ref{fig:poisson2d_relative_l2} summarizes these results and shows both the mean errors and their standard deviations, highlighting the improved accuracy and consistency achieved by SEDONet.

\begin{figure}[h!]
    \centering
    \includegraphics[width=0.60\linewidth]{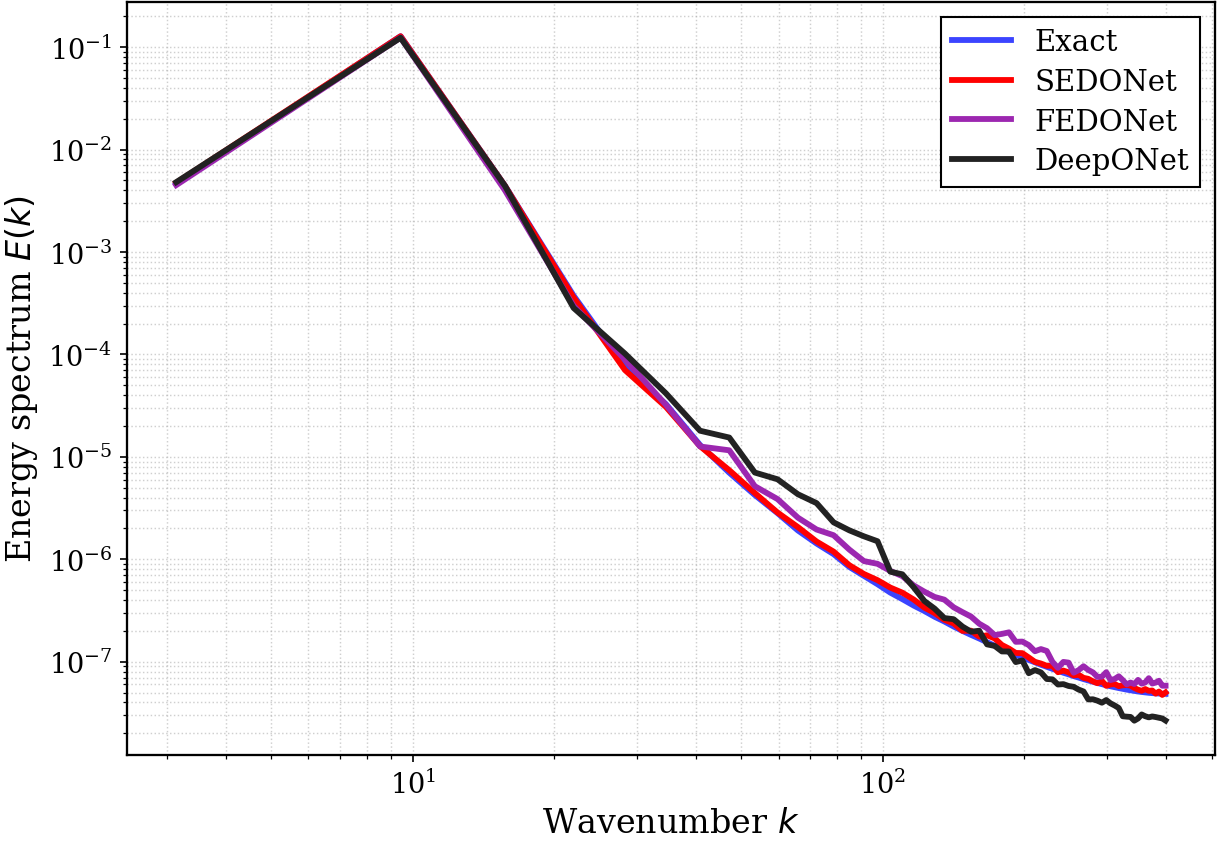}
    \caption{Energy spectra corresponding to the two-dimensional Poisson equation for the exact solution and the predictions obtained using DeepONet, FEDONet, and SEDONet. The proposed SEDONet more accurately reproduces the spectral energy distribution, particularly for the higher spatial frequencies associated with localized solution variations.}
    \label{fig:poisson2d_spectrum}
\end{figure}

Figure~\ref{fig:poisson2d_spectrum} compares the energy spectra of the reference and predicted solutions for a representative test example. All three models accurately capture the dominant low-frequency modes; however, noticeable differences emerge at intermediate and high wavenumbers. DeepONet exhibits the largest deviation from the reference spectrum, while FEDONet provides a closer approximation through Fourier-based embeddings. SEDONet shows the closest agreement with the exact spectrum across the entire wavenumber range, indicating that the proposed Chebyshev embedding more faithfully preserves fine-scale spatial structures in bounded, non-periodic domains.

\begin{figure}[h!]
    \centering
    \includegraphics[width=0.62\linewidth]{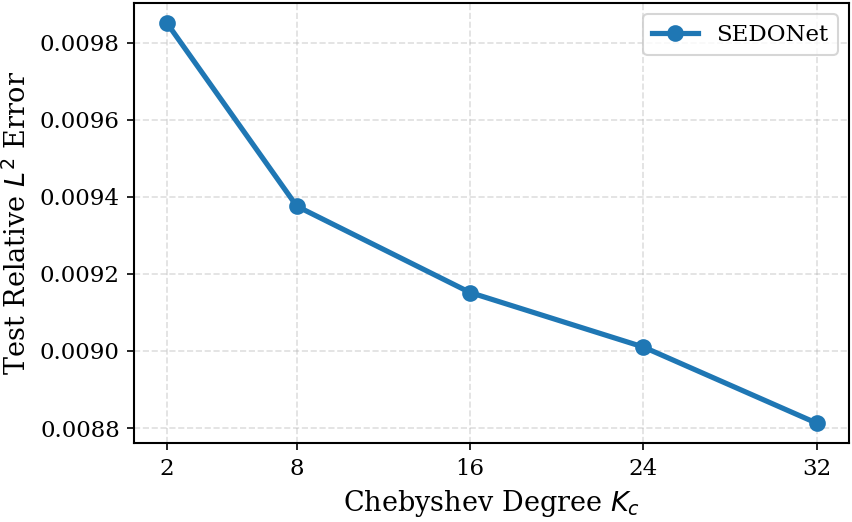}
    \caption{Sensitivity of SEDONet to the Chebyshev polynomial degree $K_c$ for the 2-D Poisson problem. The test relative $L^2$ error decreases as the polynomial degree increases, indicating that richer Chebyshev representations improve operator approximation while maintaining stable performance.}
    \label{fig:poisson_degree}
\end{figure}

To investigate the sensitivity of the proposed embedding to the polynomial degree, we perform a Chebyshev-degree ablation study for the two-dimensional Poisson problem. The polynomial degree is varied over $K_c=\{2,8,16,24,32\}$ while keeping all remaining network architectures and training hyperparameters fixed. The resulting test relative $L^2$ errors provide insight into the influence of the Chebyshev polynomial degree on predictive performance. Figure~\ref{fig:poisson_degree} shows that increasing the Chebyshev degree consistently improves the prediction accuracy, with the lowest test error obtained at $K_c=32$. This indicates that richer polynomial representations enhance the coordinate embedding for bounded elliptic problems.

\subsection{Burger's Equation}
\label{sec:burger}

We evaluate the three operator-learning architectures on the one-dimensional viscous Burger's equation defined on a bounded, non-periodic domain. The governing PDE is

\begin{equation}
    \frac{\partial u}{\partial t}(x,t)
    + u(x,t)\frac{\partial u}{\partial x}(x,t)
    = \nu \frac{\partial^{2}u}{\partial x^{2}}(x,t),
    \qquad (x,t)\in[0,1]\times[0,T],
\end{equation}

with viscosity $\nu = 0.01$ and final time $T = 0.3$. Unlike periodic benchmark settings, we impose homogeneous Dirichlet boundary conditions,

\begin{equation}
    u(0,t) = 0, \qquad u(1,t) = 0,
\end{equation}

which introduce boundary layers that are not naturally aligned with Fourier-based representations.

To generate a diverse family of smooth initial conditions, we sample a random superposition of sine modes,

\begin{equation}
    u(x,0)
    = c_1\sin(\pi x)
    + c_2\sin(2\pi x)
    + c_3\sin(3\pi x),
\end{equation}

where $c_1,c_2,c_3\sim\mathcal{N}(0,0.3^2)$. Each realization is evolved numerically using a stable explicit finite-difference scheme with upwind discretization of the advective term and centered stencils for diffusion. Trajectories exhibiting numerical instabilities (NaNs or $\|u\|_\infty>5$) are discarded and resampled. This procedure yields a final dataset of $1250$ stable space-time solutions on a grid of $N_x=N_t=100$ points.

The learning task is to approximate the nonlinear operator
\begin{equation}
    \mathcal{G}: u_0(x) \longmapsto u(x,t),
\end{equation}
mapping each initial state to its full temporal evolution. This provides a natural testbed for comparing DeepONet, FEDONet, and the proposed SEDONet under non-periodic conditions where Chebyshev embeddings are expected to offer an advantage. 

\begin{figure}[h!]
    \centering
    \includegraphics[width=0.60\linewidth]{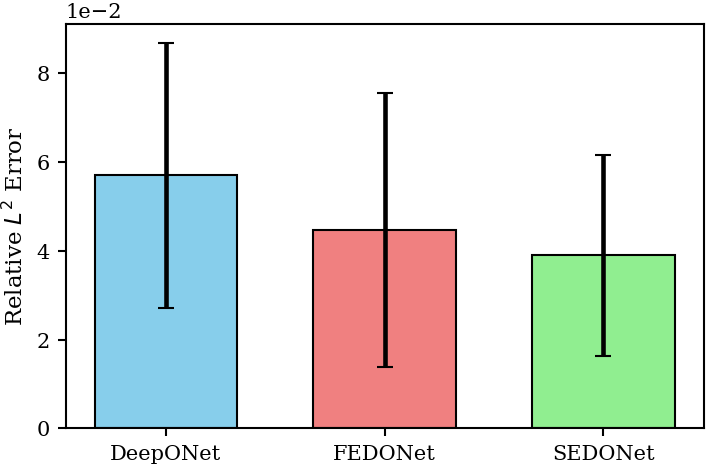}
    \caption{Relative $L^2$ error across 128 unseen Burgers' test samples for all three architectures (DeepONet, FEDONet, and SEDONet).}
    \label{fig:burgers_l2_error}
\end{figure}

Figure~\ref{fig:burgers_l2_error} reports the relative $L^2$ error over 128 unseen test samples. DeepONet exhibits the highest mean error ($5.69\%$), followed by FEDONet ($4.47\%$), while SEDONet achieves the lowest mean error of $\mathbf{3.89\%}$. This corresponds to improvements of $31.6\%$ over DeepONet and $13.0\%$ over FEDONet, demonstrating the effectiveness of the proposed Chebyshev embedding for the non-periodic Burgers' equation.

\begin{figure}[h!]
    \centering
    \includegraphics[width=0.93\linewidth]{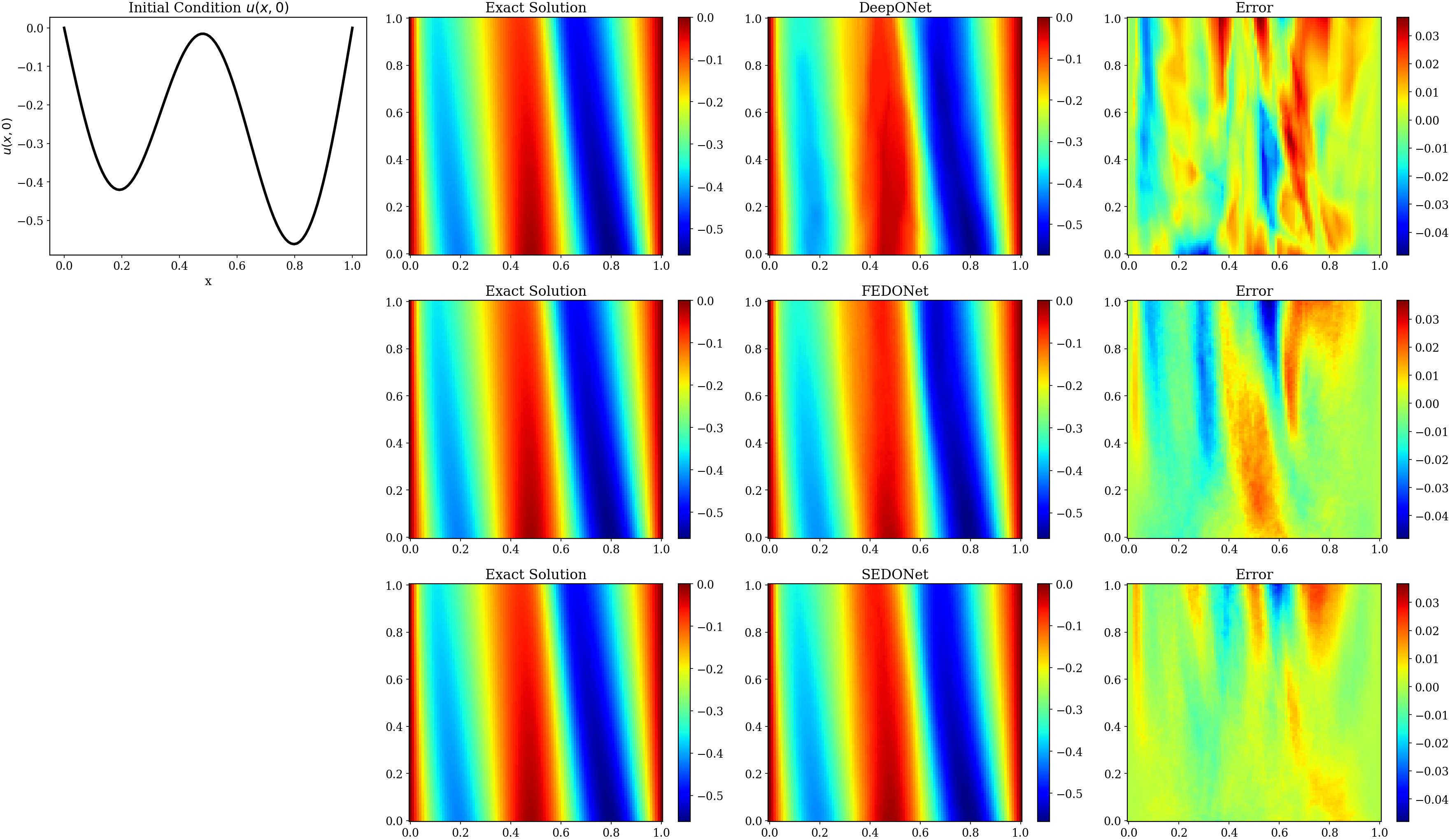}
    \caption{Best-performing Burgers' test sample: Ground truth, DeepONet, FEDONet, and SEDONet predictions with error maps.}
    \label{fig:burgers_best_sample}
\end{figure}

Figure~\ref{fig:burgers_best_sample} presents a representative test example, which shows that DeepONet reconstructs only the coarse dynamics and visibly smooths the steep gradients. While the FEDONet improves this behavior, it still struggles near sharp regions. SEDONet  provides the closest match to the ground truth, accurately tracking gradient steepening and advective transport without introducing any kind of oscillations or excessive smoothing.

\begin{figure}[H]
    \centering
    \includegraphics[width=0.93\linewidth]{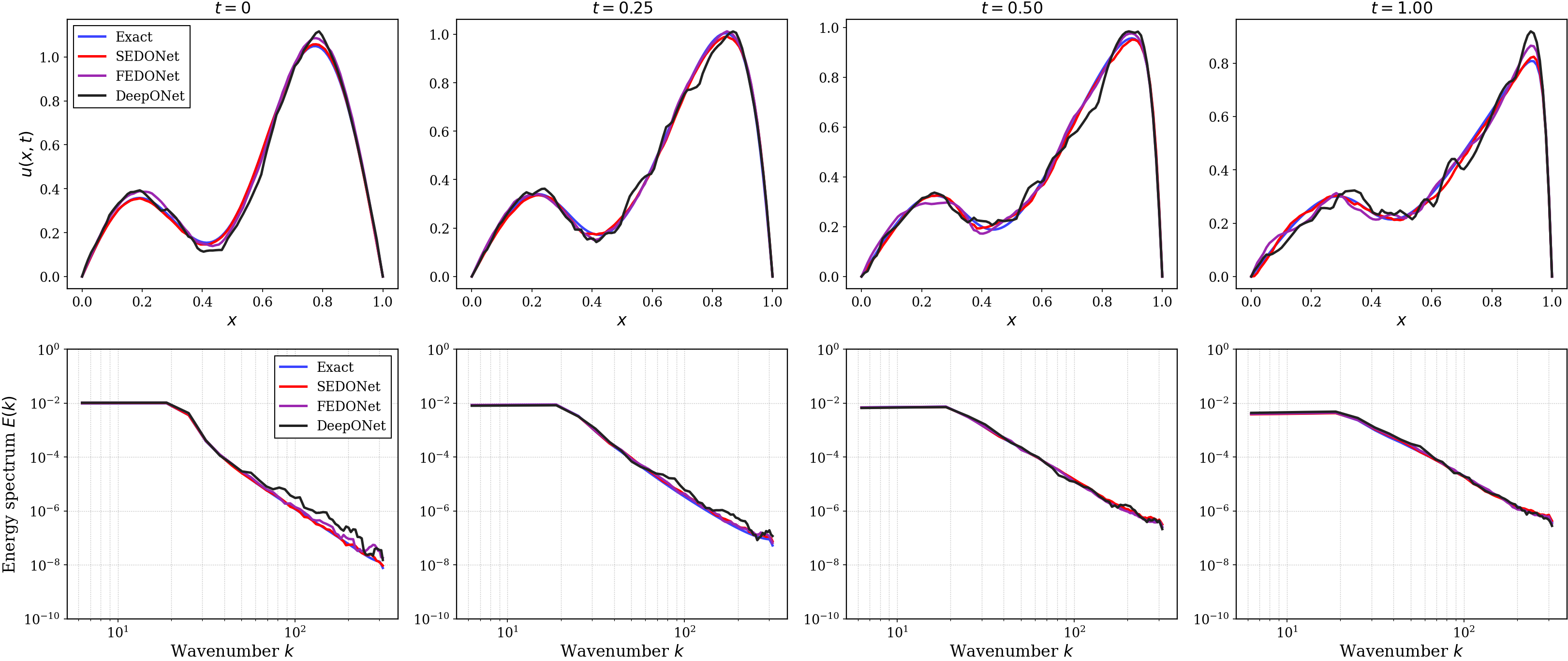}
    \caption{Energy spectrum comparison for Burgers' equation. The solution profiles (top row) and the corresponding energy spectra (bottom row) are shown at different time instances for the exact solution, DeepONet, FEDONet, and SEDONet. The proposed SEDONet most closely reproduces the reference spectral distribution throughout the temporal evolution.}
    \label{fig:burgers_spectrum}
\end{figure}

The spectral analysis shown in Figure~\ref{fig:burgers_spectrum} further highlights the performance gap, which shows that DeepONet underestimates high-wavenumber energy, indicative of its spectral bias toward smooth, low-frequency reconstructions. Also, FEDONet partially mitigates this effect through Fourier embeddings but remains limited by the non-periodic domain. SEDONet shows the closest agreement with the exact energy spectrum throughout the dissipative range, demonstrating that Chebyshev-based embeddings provide a more appropriate basis for operator learning on bounded intervals.

\begin{figure}[H]
    \centering
    \includegraphics[width=0.62\linewidth]{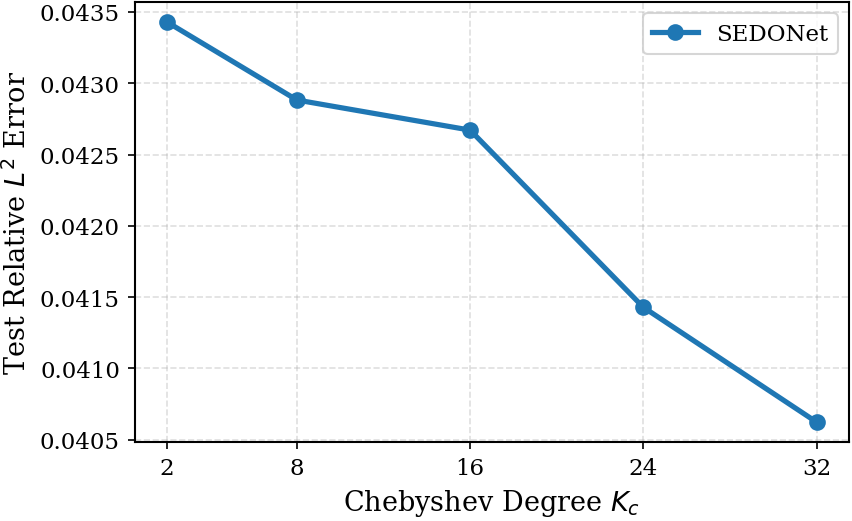}
    \caption{Sensitivity of SEDONet to the Chebyshev polynomial degree $K_c$ for the Burgers' equation. Increasing the polynomial degree consistently reduces the test relative $L^2$ error, demonstrating the benefit of higher-order Chebyshev representations for nonlinear transport problems on bounded domains.}
    \label{fig:burgers_degree}
\end{figure}

A similar Chebyshev-degree ablation study is performed for Burgers' equation to assess the sensitivity of the proposed embedding for nonlinear transport-dominated dynamics. Figure~\ref{fig:burgers_degree} shows that increasing the Chebyshev degree progressively reduces the test error before reaching a stable regime at higher polynomial orders. This indicates that richer polynomial representations improve the approximation of nonlinear transport dynamics.

Overall, the Burgers' experiments confirm that SEDONet provides a stronger inductive bias for non-periodic PDEs than both DeepONet and FEDONet. Its improved boundary handling, sharper spatial reconstructions, and more accurate spectral behavior collectively demonstrate the value of Chebyshev spectral embeddings in non-periodic operator learning.

\subsection{Advection-Diffusion Equation}
\label{sec:advection_diffusion}

We next evaluate the models on the one-dimensional advection–diffusion equation posed on a strictly non-periodic domain:
\begin{equation}
    u_t + c\,u_x = \nu\,u_{xx},
    \qquad x \in [0,1],\; t \in [0,T],
\end{equation}
subject to homogeneous Dirichlet boundary conditions,
\begin{equation}
    u(0,t) = 0, 
    \qquad 
    u(1,t) = 0.
\end{equation}
This setting reflects boundary-driven transport-diffusion processes common in physics and engineering, and provides a natural test case for architectures designed to capture non-periodic structure. The parameters are chosen as $c = 0.03$ and $\nu = 0.01$, corresponding to a diffusion-dominated regime with $\mathrm{Pe} \approx 3$. To construct a diverse set of initial states, each initial condition is formed from smooth, boundary-compatible components, including quadratic boundary-layer profiles and one or more interior Gaussian perturbations with random locations, widths, and amplitudes. All components vanish at the endpoints, ensuring that no periodic extension can represent the data and thus emphasizing the importance of non-periodic inductive biases in the trunk network.

\medskip
The spatio-temporal dataset is generated using an explicit finite-difference solver in which the advection term is approximated by a first-order upwind stencil and the diffusion term by centered second differences. The time step is selected to satisfy the parabolic stability requirement $\nu \frac{\Delta t}{\Delta x^2} \le 0.5,$ ensuring stability for all trajectories. A total of $1250$ stable simulations are produced, each consisting of an initial field $u_0(x)\in\mathbb{R}^{100}$ and the full evolution $u(t,x)\in\mathbb{R}^{100\times100}$. Any trajectory exhibiting instability or unphysical magnitudes is automatically regenerated.

\begin{figure}[h!]
    \centering
    \includegraphics[width=0.93\linewidth]{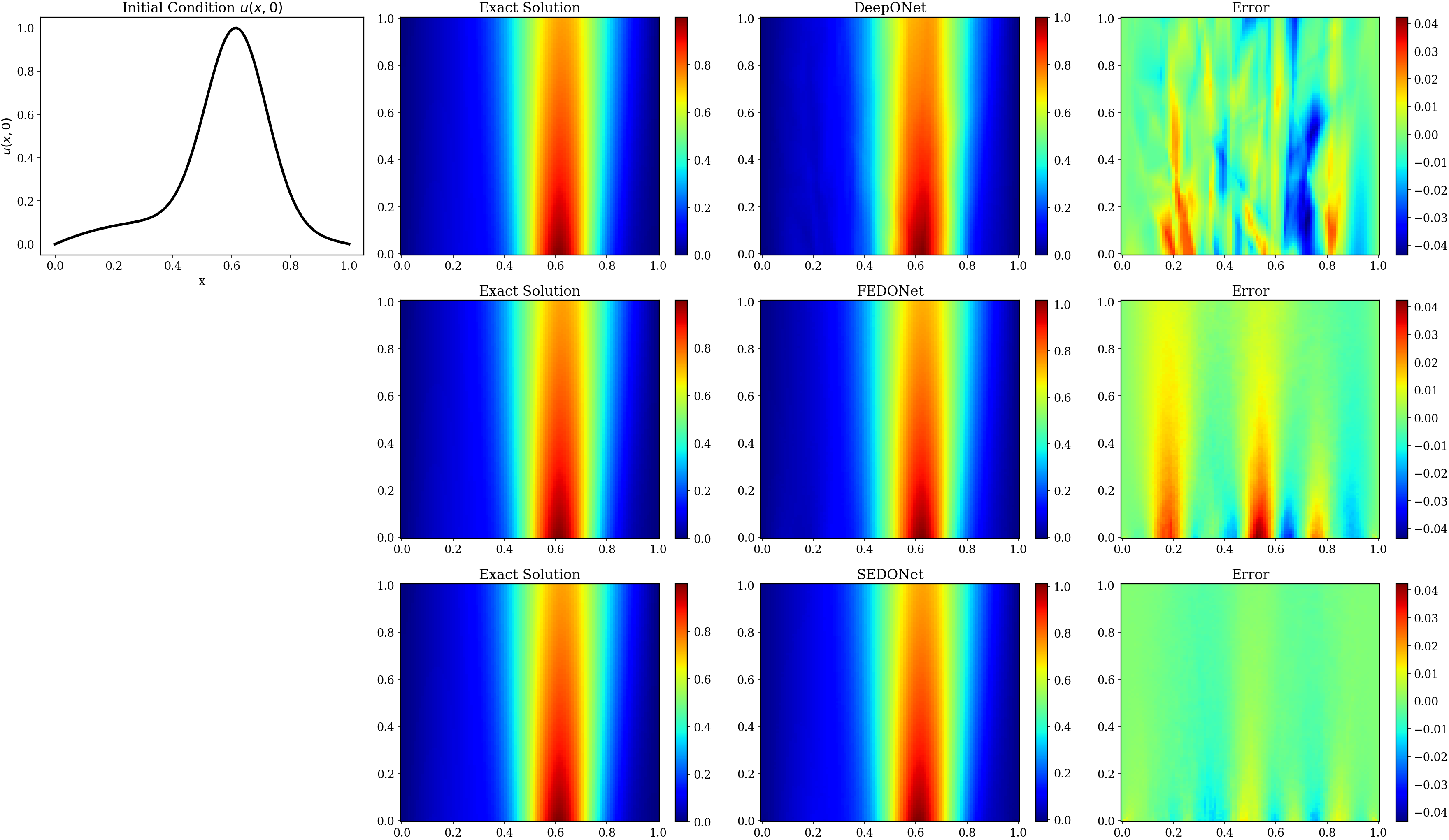}
    \caption{
        Representative test sample from the non-periodic advection-diffusion dataset.
        DeepONet exhibits smoothing and boundary distortion, FEDONet improves accuracy but retains mild periodic artifacts, and SEDONet provides the closest match to the exact solution across the full space-time domain.
    }
    \label{fig:advection-diffusion_best_sample}
\end{figure}

\medskip
Figure~\ref{fig:advection-diffusion_best_sample} shows a representative test example comparing the exact solution with predictions from DeepONet, FEDONet, and SEDONet. DeepONet captures only the coarse dynamics, oversmoothing interior gradients and producing visible distortions near the boundaries. FEDONet improves upon this behavior, but its sinusoidal inductive bias introduces mild inaccuracies close to the Dirichlet boundaries where periodic structure is inappropriate. In contrast, SEDONet achieves the closest agreement with the exact solution throughout the entire spatio-temporal domain. The Chebyshev spectral embedding aligns naturally with the non-periodic geometry, enabling sharper representation of boundary layers and more accurate tracking of diffusive smoothing.

\begin{figure}[h!]
    \centering
    \includegraphics[width=0.60\linewidth]{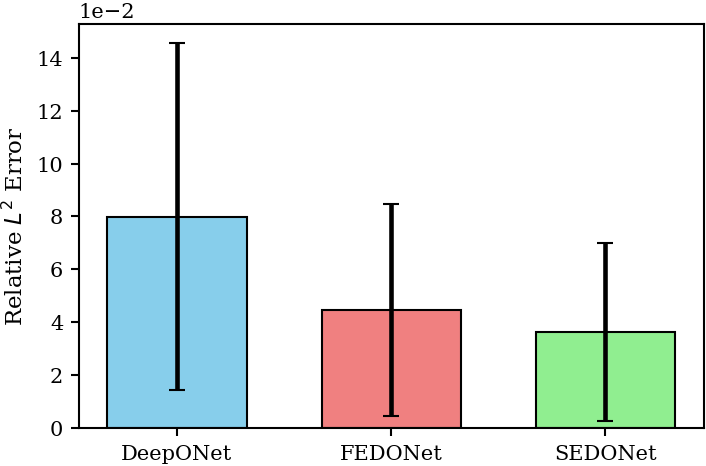}
    \caption{
        Relative $L^2$ error (mean $\pm$ std) across 250 unseen advection-diffusion test samples.
        SEDONet achieves the lowest error and improved robustness compared to both DeepONet and FEDONet.
    }
    \label{fig:advection-diffusion_best_sample_l2}
\end{figure}

\medskip
These qualitative observations are consistent with the quantitative results. Across the full unseen test set, DeepONet yields a mean relative $L^2$ error of $7.99\%$, FEDONet reduces this to $4.45\%$, and SEDONet further lowers the error to $\mathbf{3.64\%}$. This corresponds to improvements of $54.4\%$ over DeepONet and $18.2\%$ over FEDONet. The bar plot in Figure~\ref{fig:advection-diffusion_best_sample_l2}, which summarizes the mean relative $L^2$ error together with its standard deviation over the test set, further demonstrates the improved accuracy and consistency achieved by SEDONet.

\begin{figure}[h!]
    \centering
    \includegraphics[width=0.60\linewidth]{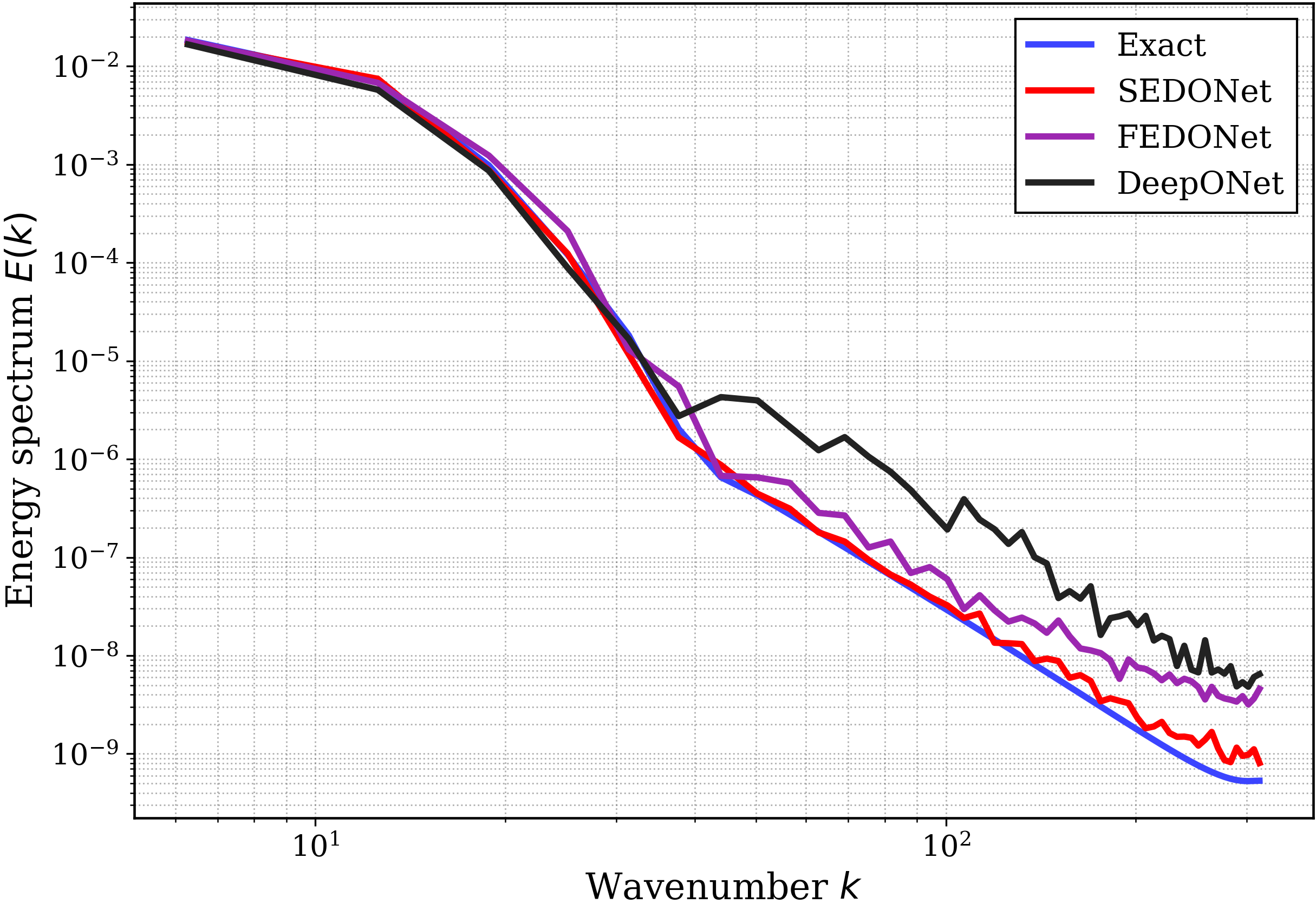}
    \caption{Energy spectra for the advection-diffusion equation comparing the exact solution with DeepONet, FEDONet, and SEDONet predictions. SEDONet most closely follows the reference spectral decay across the resolved wavenumbers.}
    \label{fig:advection-diffusion_spectrum}
\end{figure}

The corresponding energy spectra are presented in Figure~\ref{fig:advection-diffusion_spectrum}. 
While all three models accurately recover the dominant low-frequency behavior, differences become more pronounced at intermediate and high wavenumbers. In this range, DeepONet deviates noticeably from the reference spectrum and exhibits excess spectral energy over a substantial portion of the resolved modes, indicating that it does not accurately reproduce the spectral decay of the reference solution. FEDONet provides an improved spectral approximation but still shows deviations in the higher-wavenumber range. In contrast, SEDONet follows the reference spectral decay more closely across the resolved wavenumbers, demonstrating improved spectral fidelity for the non-periodic advection-diffusion problem.

\begin{figure}[h!]
    \centering
    \includegraphics[width=0.62\linewidth]{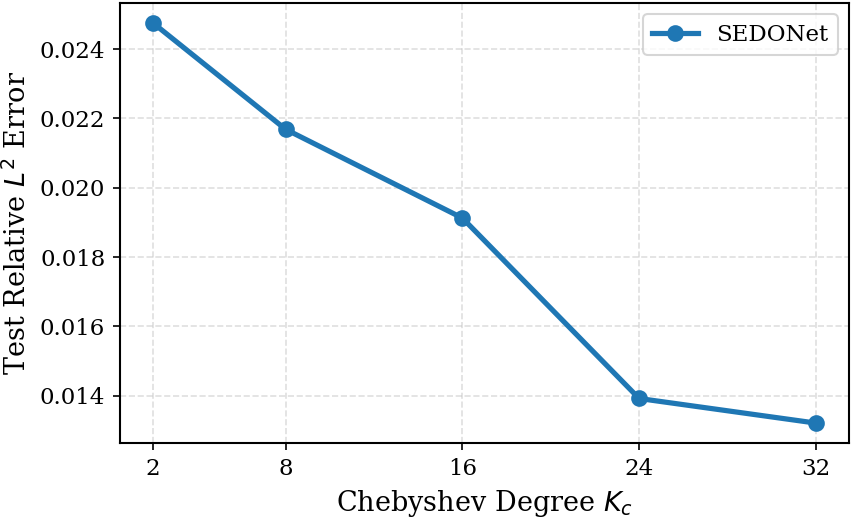}
    \caption{Sensitivity of SEDONet to the Chebyshev polynomial degree $K_c$ for the advection-diffusion equation. The results show a monotonic improvement in prediction accuracy as the polynomial degree increases, highlighting the robustness of the proposed embedding across non-periodic transport-diffusion problems.}
    \label{fig:advection_degree}
\end{figure}

We further evaluate the sensitivity of the proposed embedding to the Chebyshev polynomial degree for the advection-diffusion equation. Figure~\ref{fig:advection_degree} exhibits the same overall behavior, with prediction accuracy improving as the polynomial degree increases. The consistent trends observed across the Poisson, Burgers', and advection-diffusion problems indicate that the proposed Chebyshev embedding benefits from richer polynomial representations across different classes of non-periodic PDEs.

\medskip
Overall, this benchmark highlights that SEDONet provides a clear advantage when learning operators associated with non-periodic, boundary-constrained PDEs. The Chebyshev-based trunk offers a more suitable spectral representation than either the purely MLP-based DeepONet or the Fourier-based FEDONet, leading to improved fidelity in both physical structure and numerical accuracy.

\subsection{Lorenz-96 Chaotic System}
\label{sec:lorenz_96}

To further assess the generalization capabilities of neural operator architectures beyond PDEs, we evaluate DeepONet, FEDONet, and SEDONet on the Lorenz-96 (L96) model, a canonical benchmark in nonlinear dynamics, atmospheric modeling, and data assimilation. The governing equations are
\begin{equation}
    \frac{dx_i}{dt} = (x_{i+1}-x_{i-2})x_{i-1} - x_i + F,
    \qquad i = 1,\dots,N,
\end{equation}
with periodic boundary conditions \(x_{i \pm N} = x_i\). Here, \(x_i(t)\) denotes the \(i\)-th state variable, and the forcing parameter \(F\) controls the degree of nonlinearity. We set \(N = 40\) and \(F = 4.0\), producing smooth but strongly nonlinear dynamics with coherent wave-like propagation. The system is integrated using a fourth-order Runge-Kutta (RK4) method with time step \(\Delta t = 0.01\). Initial conditions are generated by perturbing the equilibrium state \(x_i = F\) with small Gaussian noise, $
    \mathbf{x}_0 = F\mathbf{1} + \epsilon\,\mathcal{N}(0, I),
    \quad \epsilon = 10^{-3}.
$ Each trajectory is simulated for \(15\) seconds, where the initial \(10\) seconds are discarded to remove transients. The remaining \(5\) seconds form a sequence of \(501\) snapshots. Repeating this for \(10{,}000\) initial conditions produces a large ensemble of spatio-temporal trajectories with shape \((10{,}000,\; 501,\; 40)\).

\begin{figure}[h!]
    \centering
    \includegraphics[width=0.93\linewidth]{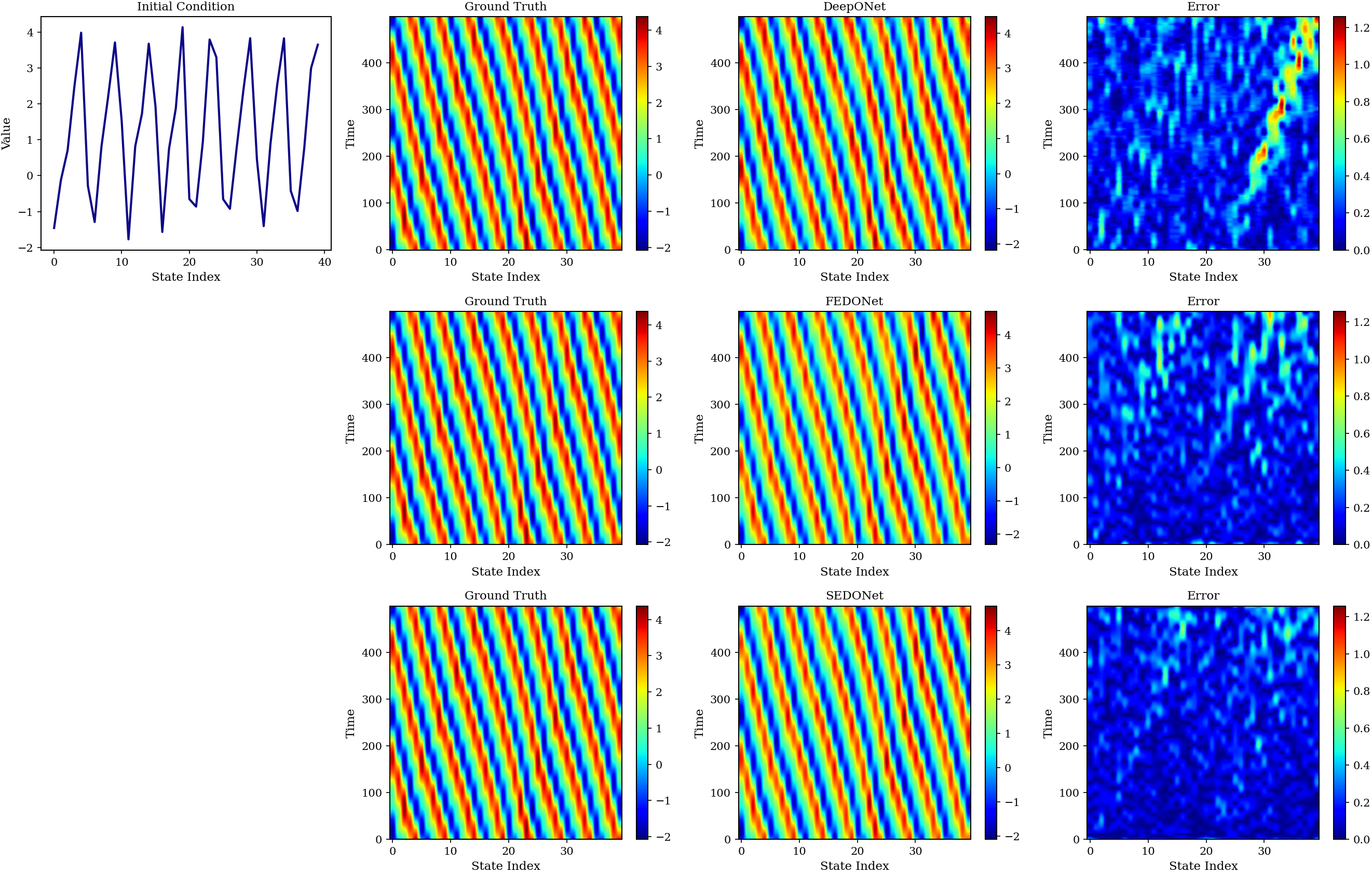}
    \caption{
        Spatio-temporal reconstruction of representative Lorenz-96 trajectories: ground truth, DeepONet, FEDONet, and SEDONet predictions, together with absolute error fields.
    }
    \label{fig:l96_contours}
\end{figure}

\medskip
Figure~\ref{fig:l96_contours} shows representative test reconstructions. Each row includes the initial condition, the exact spatio-temporal evolution, predictions from DeepONet, FEDONet, and SEDONet, followed by the absolute error fields. All three models reproduce the dominant diagonal wave structures and maintain the phase velocity over the full prediction horizon. Vanilla DeepONet, however, shows mild phase drift and slight amplitude damping, especially in regions with sharper gradients. FEDONet alleviates some of these issues due to the Fourier-based positional encoding, though the improvement is modest because the Lorenz-96 attractor is inherently smooth.

\begin{figure}[h!]
    \centering
    \includegraphics[width=0.60\linewidth]{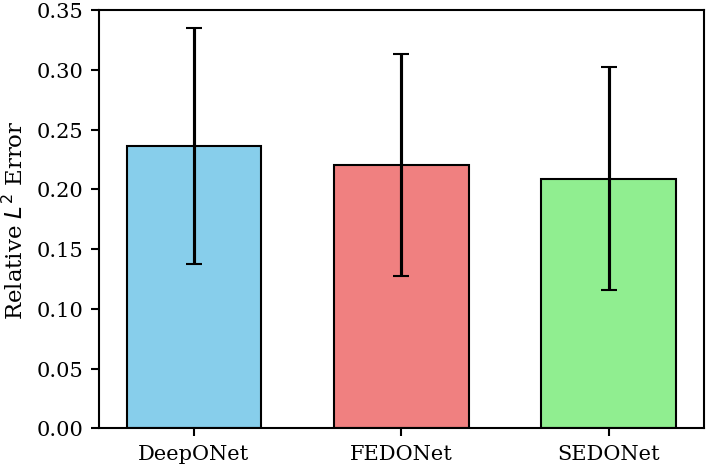}
    \caption{
        Relative $L^2$ error (mean $\pm$ std) across 2000 unseen Lorenz-96 trajectories.
        SEDONet achieves both the lowest average error and the lowest variance, demonstrating improved stability and robustness compared to DeepONet and FEDONet.
    }
    \label{fig:l96_contours_l2}
\end{figure}

\medskip
SEDONet provides the most accurate long-horizon reconstruction. Its Chebyshev-based spectral trunk better regulates amplitude and reduces cumulative error, resulting in cleaner residual fields and smaller drift in regions where DeepONet and FEDONet deviate. This behavior is visible throughout the full spatio-temporal domain. These qualitative observations are further supported by the quantitative evaluation summarized in Figure~\ref{fig:l96_contours_l2}. The bar plot displays the mean relative $L^2$ error and associated standard deviation across 2000 unseen trajectories. DeepONet attains an average error of approximately $23.6\%$, FEDONet reduces this slightly to around $22.0\%$, and SEDONet further lowers the error to about $\mathbf{20.9\%}$. In addition to having the lowest mean error, SEDONet also exhibits reduced variance, indicating more consistent performance across different initial conditions. Although the absolute improvements are smaller than those observed for elliptic or convection-diffusion PDEs, the trend remains robust, SEDONet achieves the best stability and predictive accuracy among the three architectures on long-horizon chaotic dynamics.

\begin{figure}[h!]
    \centering
    \includegraphics[width=0.59\linewidth]{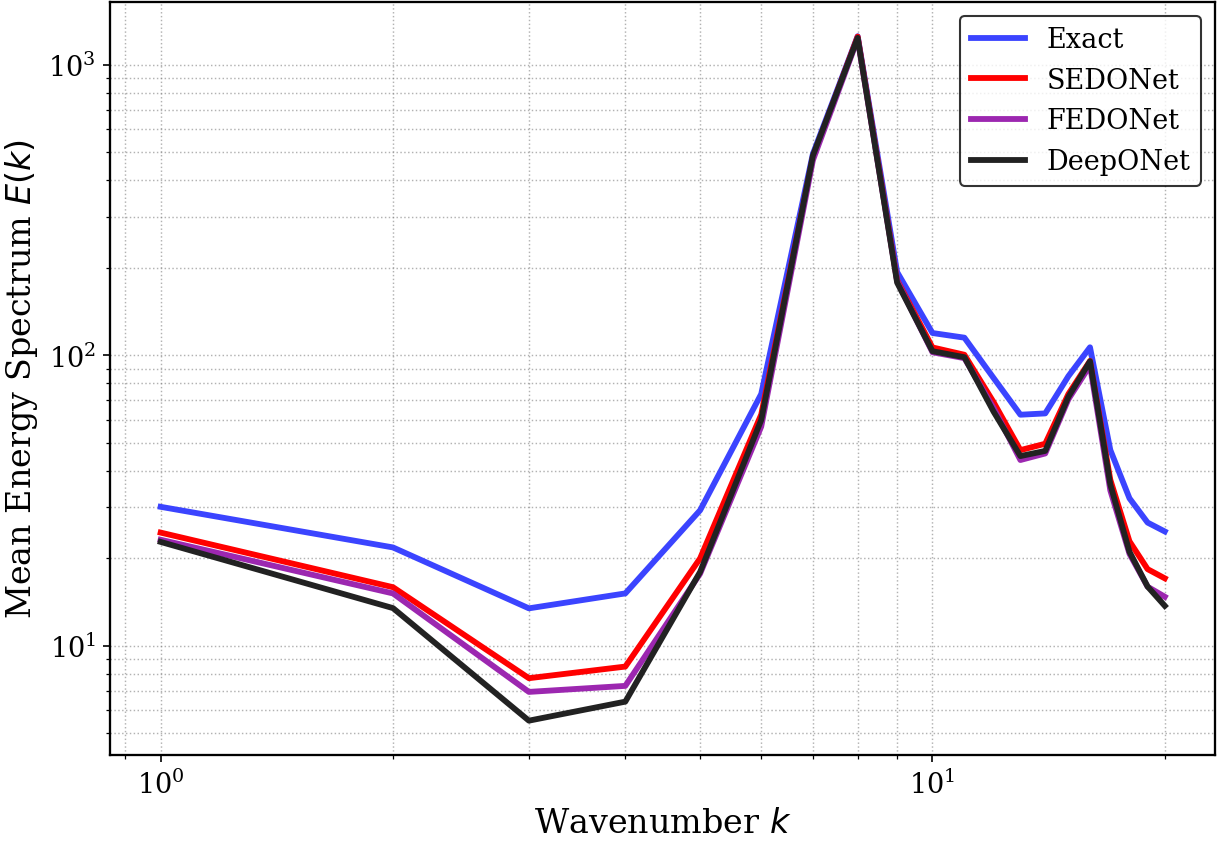}
    \caption{Comparison of the spectral distributions for the Lorenz-96 system obtained using DeepONet, FEDONet, and SEDONet. The proposed SEDONet captures the energy across the resolved modes, indicating an improved representation of the multiscale dynamics.}
    \label{fig:l96_spectrum}
\end{figure}

Figure~\ref{fig:l96_spectrum} presents the energy spectrum for a representative Lorenz-96 trajectory. DeepONet captures the dominant large-scale modes but progressively loses energy in the higher-frequency components. FEDONet improves the spectral representation through Fourier embeddings, whereas SEDONet most closely follows the reference spectrum over the full range of resolved modes. This indicates that the proposed Chebyshev embedding better preserves the multiscale characteristics of the learned operator.

\medskip
Overall, these results indicate that even for smooth, quasi-periodic systems like Lorenz-96, where fully connected trunks already perform reasonably well, the Chebyshev spectral embeddings of SEDONet still yield measurable benefits. The architecture improves stability, reduces phase and amplitude drift, and consistently achieves lower reconstruction error across thousands of independent trajectories.

\subsection{Allen-Cahn Equation}
\label{sec:allen_cahn}

We further benchmark the three operator-learning architectures on the one-dimensional Allen-Cahn equation, a nonlinear reaction-diffusion model describing phase separation and interface motion. The governing PDE is
\begin{equation}
    \frac{\partial u}{\partial t}
    = \epsilon\, u_{xx} - 5u^3 + 5u,
    \qquad x \in [-1,1], \quad t \in [0,1],
\end{equation}
where \(u(x,t)\) is the phase-field variable and \(\epsilon = 10^{-4}\) controls the interface thickness. Periodic boundary conditions are imposed on both \(u\) and \(u_x\), consistent with phase-field models on closed spatial domains. A dataset of \(10{,}000\) trajectories is generated using an explicit Euler integrator with spatial resolution \(\Delta x = 0.01\) and time step \(\Delta t = 0.005\). The initial condition for each trajectory is constructed as
\begin{equation}
    s(x) =
    \sum_{k=1}^{3}
    \Big[
         a_k\, x^{2k} \cos(k\pi x)
        + b_k\, x^{2k} \sin(k\pi x)
    \Big],
\end{equation}
where \(a_k, b_k \sim \mathcal{U}(0,1)\). This yields sharply varying, multiscale profiles with strong spatial gradients, precisely the type of regime where spectral biases play a significant role. Each sample consists of the initial field \(s(x) \in \mathbb{R}^{200}\) and the full spatio-temporal solution \(u(x,t) \in \mathbb{R}^{200\times 200}\). The learning task is to approximate the nonlinear solution operator $\mathcal{G}: s(x) \mapsto u(x,t).$

\begin{figure}[h!]
    \centering
    \includegraphics[width=0.93\linewidth]{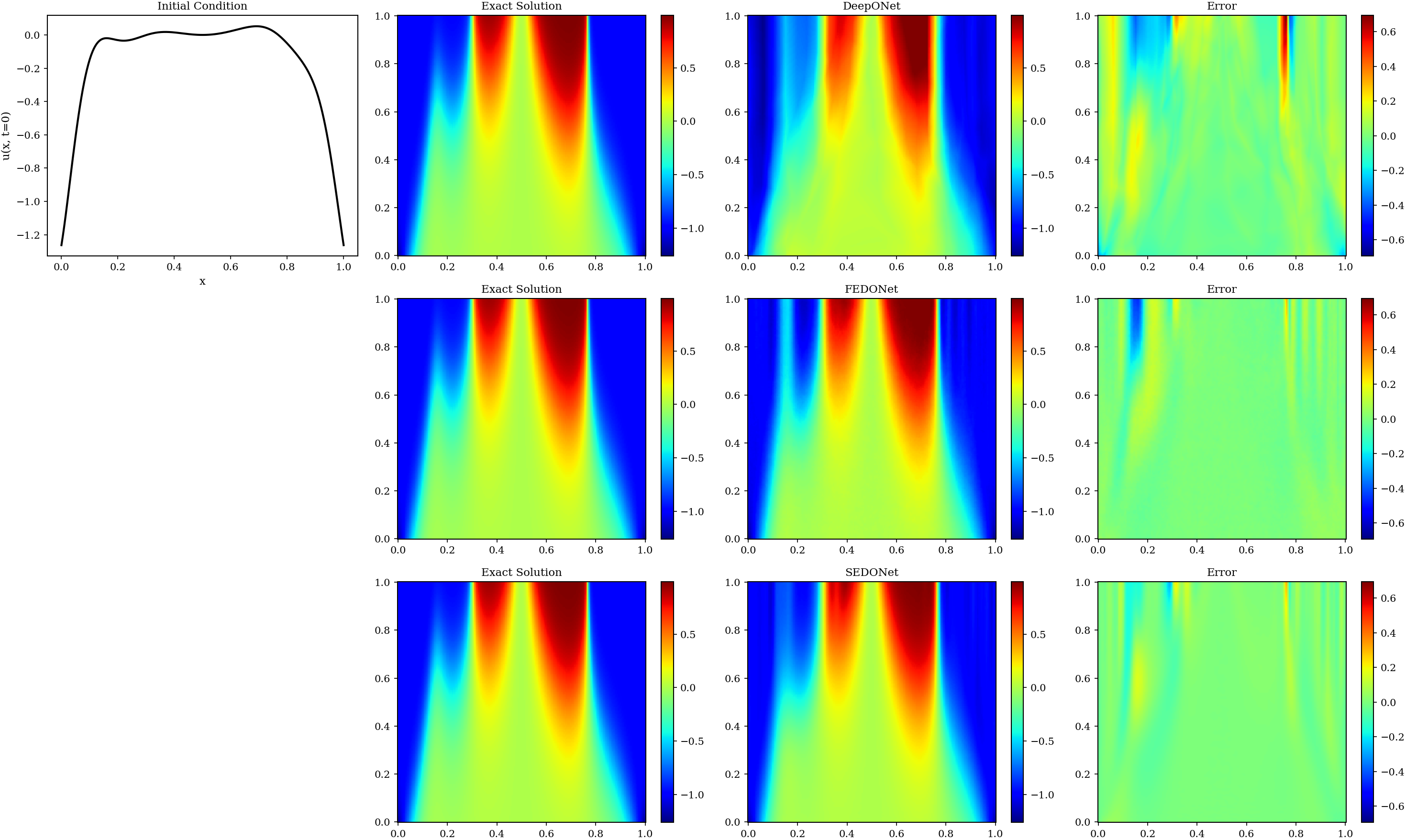}
    \caption{
        Spatio-temporal reconstruction for a representative Allen-Cahn test sample: ground truth, DeepONet, FEDONet, and SEDONet predictions, together with pointwise error fields.
    }
    \label{fig:AC_best}
\end{figure}

\medskip
Figure~\ref{fig:AC_best} presents a representative test example for the Allen-Cahn equation. DeepONet captures the overall phase evolution but exhibits noticeable smoothing of the interface regions, resulting in larger reconstruction errors near steep transitions. FEDONet substantially improves the reconstruction by better preserving the interface dynamics and reducing the pointwise residuals. SEDONet provides the closest agreement with the reference solution, exhibiting slightly sharper interface resolution and lower residual magnitudes throughout the spatio-temporal domain. Overall, the qualitative comparison is consistent with the quantitative results, showing that both spectral embedding approaches significantly outperform the baseline DeepONet, with SEDONet providing the most accurate reconstruction for this benchmark.

\begin{figure}[h!]
    \centering
    \includegraphics[width=0.6\linewidth]{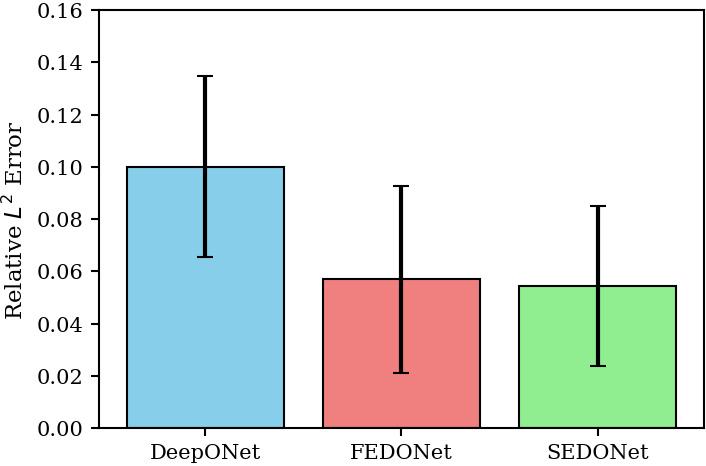}
    \caption{
        Relative $L^2$ error (mean $\pm$ std) across 250 unseen Allen--Cahn test samples for DeepONet, FEDONet, and SEDONet.
        SEDONet achieves the lowest error with reduced variance across the ensemble.
    }
    \label{fig:AC_beste_l2}
\end{figure}

\medskip
These qualitative observations are further supported by the quantitative results summarized in Figure~\ref{fig:AC_beste_l2}. DeepONet attains a mean relative $L^2$ error of $10.01\%$, while the two spectral embedding approaches substantially improve the prediction accuracy. FEDONet reduces the error to $5.70\%$, and SEDONet further lowers it to $\mathbf{5.44\%}$, corresponding to a modest but consistent improvement over FEDONet. In addition, SEDONet exhibits slightly lower variability across the test set, indicating more consistent predictive performance. These results demonstrate that the proposed Chebyshev embedding remains competitive on this periodic benchmark while providing the best overall accuracy among the three architectures.

\begin{figure}[h!]
    \centering
    \includegraphics[width=0.6\linewidth]{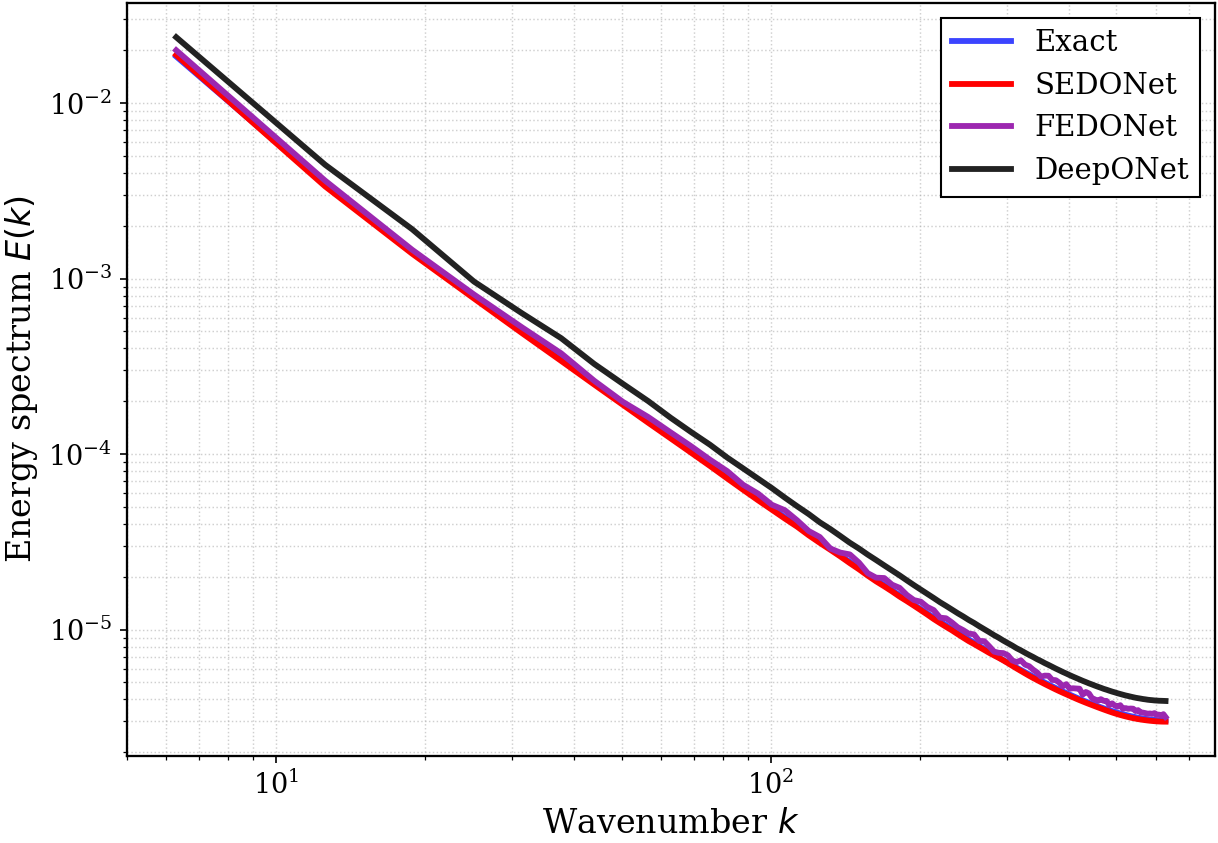}
    \caption{Comparison of the energy spectra for the Allen--Cahn equation obtained using DeepONet, FEDONet, and SEDONet. The spectral distributions provide a complementary assessment of each model's ability to preserve the frequency content of the solution across the resolved modes.}
    \label{fig:AC_spectrum}
\end{figure}

Figure~\ref{fig:AC_spectrum} presents the corresponding energy spectra for the Allen-Cahn equation. Overall, all three neural operator architectures accurately reproduce the spectral distribution of the reference solution, with only small differences observed at intermediate and high wavenumbers. Compared with the baseline DeepONet, both spectral embedding approaches more closely follow the reference spectrum, indicating an improved ability to preserve the multiscale characteristics of the evolving phase-field solution. This behavior is consistent with the quantitative results, where FEDONet and SEDONet achieve comparable predictive performance while both outperform the baseline DeepONet.

\medskip
Overall, the Allen-Cahn benchmark demonstrates that spectral embeddings are beneficial for learning nonlinear phase-field dynamics. While the improvements obtained with the proposed Chebyshev embedding are more modest than those observed for the non-periodic PDE benchmarks, SEDONet remains competitive and achieves the lowest overall prediction error among the three architectures. These results further suggest that the proposed Chebyshev embedding maintains robust predictive performance even when applied to periodic problems, while offering its greatest advantage for bounded, non-periodic operator-learning tasks.
\subsection{Darcy Flow in a Rectangular Domain}
\label{sec:darcy_flow}

The Darcy flow benchmark is defined on the rectangular domain $\Omega=[0,1]^2$ with homogeneous Dirichlet boundary conditions. The objective is to learn the nonlinear operator mapping the heterogeneous permeability field $a(x,y)$ to the corresponding pressure solution $u(x,y)$,
\begin{equation}
\mathcal{G}: a(x,y) \mapsto u(x,y).
\end{equation}
The governing steady-state elliptic PDE is
\begin{equation}
-\nabla \cdot \left(a(x,y)\nabla u(x,y)\right)=f(x,y), \qquad (x,y)\in\Omega,
\end{equation}
where $a(x,y)$ denotes the permeability coefficient, $u(x,y)$ is the pressure field, and $f(x,y)$ is the forcing term. The permeability fields are sampled from a Gaussian random field, and the corresponding pressure solutions are computed numerically on a fine-resolution mesh of size $421\times421$, following the standard Darcy benchmark introduced in~\cite{li2021fourierneuraloperatorparametric}. Consistent with previous neural operator studies, the dataset is downsampled using a spatial reduction factor of $r=4$, yielding an effective resolution of $106\times106$.

Unlike the transient PDE benchmarks considered above, the Darcy problem represents a steady-state elliptic system in which the main challenge lies in propagating localized permeability variations through a global elliptic operator. This provides a useful test case for evaluating whether different trunk embeddings can represent non-periodic spatial heterogeneity. The branch architecture and optimization hyperparameters are kept identical across DeepONet, FEDONet, and SEDONet to ensure a controlled comparison. The only difference among the three models is the trunk representation: DeepONet uses a standard coordinate-input MLP, FEDONet uses Fourier feature embeddings, and SEDONet uses the proposed Chebyshev polynomial embedding designed for bounded, non-periodic domains.

\begin{figure}[h!]
    \centering
    \includegraphics[width=0.93\linewidth]{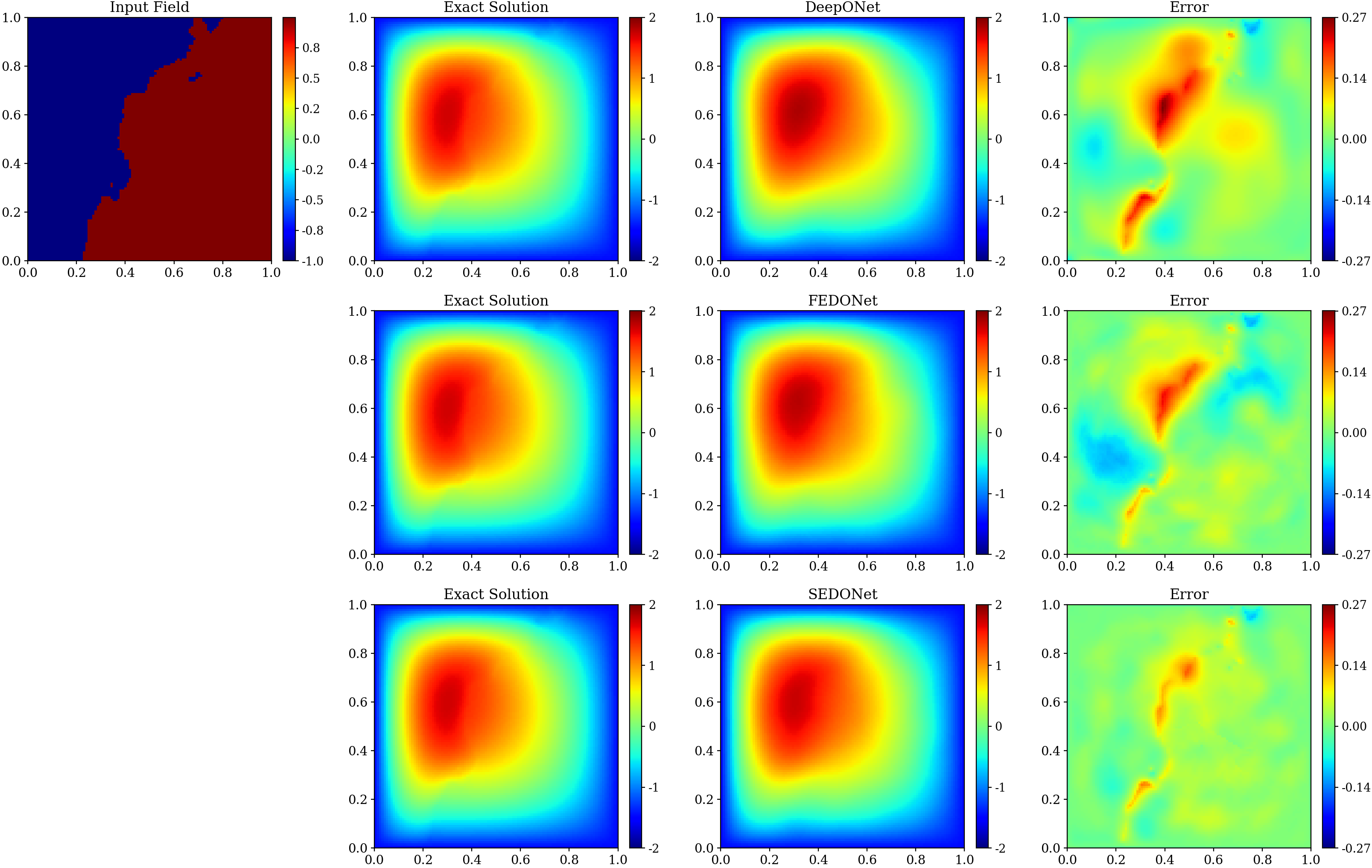}
    \caption{Prediction comparison for the Darcy flow benchmark showing the input permeability field $a(x,y)$, the ground-truth pressure solution $u(x,y)$, and the corresponding predictions and error maps for DeepONet, FEDONet, and SEDONet.}
    \label{fig:darcy_best}
\end{figure}

The permeability field contains localized discontinuities and multiscale spatial heterogeneity, making this benchmark challenging for global operator learning. Fourier embeddings provide a useful spectral representation, but their periodic inductive bias is not naturally aligned with bounded elliptic problems. In contrast, the Chebyshev polynomial basis forms an orthogonal representation on finite intervals, allowing SEDONet to represent localized spatial variations while respecting the non-periodic nature of the domain. Figure~\ref{fig:darcy_best} presents qualitative comparisons for the Darcy benchmark. All three neural operators recover the dominant pressure distribution, but the residual fields reveal differences in local accuracy. DeepONet produces larger spatially distributed errors, while FEDONet reduces these residuals through spectral coordinate embeddings. SEDONet further decreases the error magnitude, particularly near regions of rapidly varying permeability, indicating that the Chebyshev embedding provides a suitable inductive bias for bounded elliptic operators.

\begin{figure}[h!]
    \centering
    \includegraphics[width=0.60\linewidth]{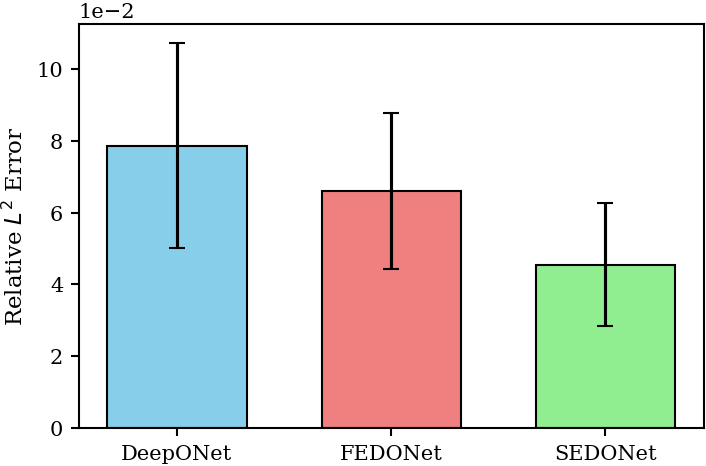}
    \caption{Test relative $L^2$ error for the Darcy flow benchmark. Error bars indicate the mean and standard deviation over the test set. SEDONet achieves the lowest prediction error among the three architectures.}
    \label{fig:darcy_l2}
\end{figure}

The quantitative performance comparison is summarized in Figure~\ref{fig:darcy_l2}. DeepONet achieves a mean relative $L^2$ error of $7.86\%$, while FEDONet reduces the error to $6.59\%$. SEDONet further decreases the error to $\mathbf{4.55\%}$, corresponding to improvements of $42.1\%$ over DeepONet and $30.9\%$ over FEDONet. These results demonstrate that the proposed Chebyshev embedding provides a more accurate approximation of the Darcy flow operator on bounded, non-periodic domains.

\begin{figure}[h!]
    \centering
    \includegraphics[width=0.62\linewidth]{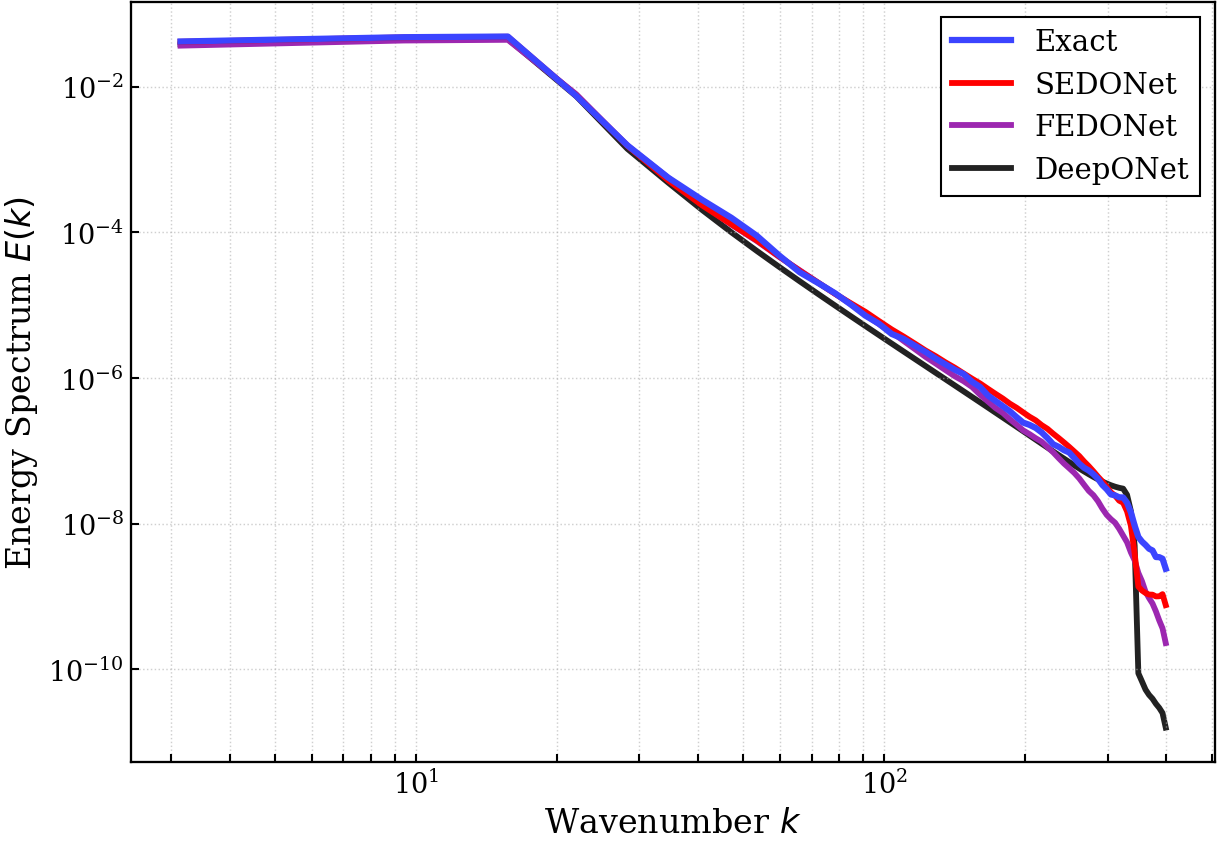}
    \caption{Comparison of the energy spectra for the Darcy flow problem. The spectra illustrate how each model represents the distribution of solution energy across spatial scales.}
    \label{fig:darcy_spectrum}
\end{figure}

Figure~\ref{fig:darcy_spectrum} presents the energy spectra for the Darcy flow benchmark. Although all three architectures recover the dominant low-frequency behavior, differences become more apparent at intermediate and high wavenumbers. SEDONet more closely follows the reference spectrum in these ranges, indicating improved representation of fine-scale spatial variations in the elliptic solution field.

These results demonstrate that replacing Fourier embeddings with Chebyshev polynomial embeddings provides a clear advantage for bounded, non-periodic elliptic operator learning. The Darcy benchmark therefore complements the transient PDE experiments by showing that the proposed SEDONet architecture also improves operator approximation accuracy in steady-state problems where no temporal dynamics are present.

\subsection{Generalization to Other Neural Operator Architectures}
\label{subsec:model_generalization}
To investigate whether the proposed Chebyshev embedding is specific to the DeepONet architecture or can be generalized to other neural operators, we additionally evaluated the proposed embedding within the Fourier Neural Operator (FNO). Unlike SEDONet, which introduces Chebyshev embeddings into the trunk network of DeepONet, the FNO implementation replaces the coordinate representation used by the lifting layer with the proposed Chebyshev embedding while leaving the remaining network architecture unchanged. This experiment aims to assess whether the benefits of Chebyshev coordinate representations extend beyond the branch-trunk decomposition employed by DeepONet.

\begin{table}[ht]
\centering
\caption{Generalization of the proposed Chebyshev embedding to the Fourier Neural Operator (FNO). Mean relative $L^2$ errors are reported over the test set.}
\label{tab:fno}
\begin{tabular}{lccc}
\toprule
Problem & FNO & FNO + Chebyshev & Improvement (\%)\\
\midrule
Poisson &
0.002321 $\pm$ 0.000539 &
\textbf{0.001279 $\pm$ 0.000369} &
44.9 \\

Burgers &
0.002551 $\pm$ 0.001951 &
\textbf{0.001490 $\pm$ 0.001067} &
41.6 \\

Advection-Diffusion &
0.003702 $\pm$ 0.004506 &
\textbf{0.002302 $\pm$ 0.003094} &
37.8 \\
\bottomrule
\end{tabular}
\end{table}

Table~\ref{tab:fno} demonstrates that the proposed Chebyshev embedding consistently improves the predictive accuracy of FNO across all three benchmark problems. Since the branch-trunk decomposition is unique to DeepONet and is absent in FNO, these results indicate that the benefits of the proposed embedding arise from the coordinate representation itself rather than from a specific network architecture. This observation suggests that Chebyshev-based coordinate embeddings constitute a general spectral representation strategy that can be incorporated into multiple neural operator frameworks for bounded, non-periodic problems.
\section{Summary}
\label{Summary}

This work introduced the Spectral-Embedded DeepONet (SEDONet), a Chebyshev-enhanced operator-learning architecture tailored to PDEs on bounded and non-periodic domains. By replacing the standard coordinate-input trunk with a fixed tensor-product Chebyshev dictionary, SEDONet incorporates a non-periodic spectral prior that strengthens the representation of boundary layers, steep gradients, and complex spatio-temporal solution structures. The branch network remains unchanged, preserving the original DeepONet framework without introducing additional trainable parameters.

The proposed approach was evaluated on six benchmark problems spanning elliptic, hyperbolic, advection--diffusion, phase-field, chaotic, and Darcy flow systems. As summarized in Table~\ref{tab:relL2}, SEDONet consistently achieves the lowest or statistically comparable relative $L^2$ errors across all benchmark problems, outperforming the baseline DeepONet and providing meaningful improvements over the Fourier-Embedded DeepONet (FEDONet). On canonical non-periodic problems such as the 2-D Poisson equation, Burgers' equation, the advection--diffusion equation, and Darcy flow, the proposed Chebyshev embedding produces more accurate solution reconstructions and improved representation of boundary and interior solution features. For the periodic benchmarks, namely the Lorenz--96 system and the Allen--Cahn equation, SEDONet remains competitive while maintaining comparable or lower prediction errors than the baseline DeepONet.

Beyond pointwise reconstruction accuracy, the accompanying energy spectrum analyses demonstrate that SEDONet more faithfully reproduces the spectral distribution of the reference solutions, particularly for the non-periodic PDE benchmarks where preserving intermediate- and high-frequency components is essential for accurately representing fine-scale solution structures. Furthermore, the additional Chebyshev degree ablation study confirms the robustness of the proposed embedding with respect to the polynomial order, while the extension of the proposed embedding to the Fourier Neural Operator (FNO) demonstrates that the underlying Chebyshev coordinate representation is not limited to the DeepONet architecture and can be effectively incorporated into other neural operator frameworks.

\begin{table}[ht]
  \centering
  \scriptsize
  \begin{threeparttable}
    \caption{Relative $L^2$ error (\%, mean $\pm$ std) for each benchmark.\label{tab:relL2}}
    \begin{tabularx}{\linewidth}{l
        >{\centering\arraybackslash}X
        >{\centering\arraybackslash}X
        >{\centering\arraybackslash}X}
      \toprule
      \textbf{Dataset} &
      \textbf{DeepONet} &
      \textbf{FEDONet} &
      \textbf{SEDONet} \\
      \midrule
      2-D Poisson (elliptic)
        & $1.39 \pm 0.71$
        & $1.10 \pm 0.52$
        & $\mathbf{0.97 \pm 0.50}$ \\
        
      1-D Burgers
        & $5.69 \pm 2.98$
        & $4.47 \pm 3.08$
        & $\mathbf{3.89 \pm 2.26}$ \\

      1-D Advection-Diffusion
        & $7.99 \pm 6.55$
        & $4.45 \pm 4.05$
        & $\mathbf{3.64 \pm 3.35}$ \\
      
      Lorenz-96 (chaotic ODE)
        & $23.63 \pm 9.88$
        & $22.03 \pm 9.26$
        & $\mathbf{20.90 \pm 9.33}$ \\

      Allen-Cahn (phase-field)
        & $10.01 \pm 3.46$
        & $5.70 \pm 3.57$
        & $\mathbf{5.44 \pm 3.05}$ \\

      2-D Darcy Flow
        & $7.86 \pm 2.85$
        & $6.59 \pm 2.17$
        & $\mathbf{4.55 \pm 1.70}$ \\        
      \bottomrule
    \end{tabularx}
  \end{threeparttable}
\end{table}

Overall, the benchmark results indicate that the proposed Chebyshev embedding provides a particularly suitable inductive bias for bounded, non-periodic domains, where the polynomial basis naturally aligns with the underlying boundary conditions. Although the most significant improvements are observed for non-periodic PDEs, SEDONet remains competitive on periodic benchmark problems without degrading predictive performance. Furthermore, the proposed embedding preserves the original DeepONet architecture and introduces no additional trainable parameters, while providing a richer spectral representation of the coordinate space. The presented Chebyshev-degree ablation studies also indicate that the optimal polynomial degree may depend on the underlying problem and spatial resolution. While the proposed embedding has been successfully extended to the Fourier Neural Operator (FNO), its application to other neural operator architectures, more complex geometries, and broader classes of PDEs remains an important direction for future work. Consequently, SEDONet provides a simple, modular, and parameter-efficient enhancement to DeepONet that bridges classical polynomial spectral methods with modern neural operator architectures for learning nonlinear operators arising in scientific computing.

\section{Future Work}
\label{Futurework}

The proposed Spectral-Embedded DeepONet (SEDONet) provides a flexible framework for incorporating Chebyshev polynomial representations into neural operator learning, particularly for bounded, non-periodic domains. Although the present work demonstrates consistent improvements across a diverse collection of benchmark problems, several promising directions remain for future research. A natural extension is the development of adaptive spectral embeddings within the trunk network. In the current formulation, the Chebyshev dictionary is fixed a priori. Allowing the polynomial degree, scaling factors, or spectral weights to be learned jointly with the network parameters may enable the coordinate representation to adapt automatically to the local smoothness and complexity of the underlying solution. Such adaptive spectral embeddings could be particularly beneficial for problems involving moving interfaces, sharp gradients, or highly heterogeneous material properties.

Another promising direction is the construction of hybrid spectral trunks that combine Chebyshev polynomials with complementary basis functions, such as Legendre polynomials, Gegenbauer polynomials, or localized wavelets, to better capture solution behavior exhibiting mixed regularity. Extending the proposed framework to irregular geometries through coordinate transformations or graph-based representations would further broaden its applicability to practical engineering and geophysical problems. Furthermore, the present study demonstrates that the proposed Chebyshev embedding can be successfully incorporated into the Fourier Neural Operator (FNO). Extending this idea to other neural operator architectures represents an important avenue for future work and may provide a general spectral embedding framework applicable across a broad range of operator-learning models.

Finally, integrating SEDONet with physics-informed training strategies, uncertainty quantification, or data assimilation techniques offers another promising research direction. Embedding governing equations, conservation constraints, or probabilistic priors into the learning process may further improve robustness, particularly for scientific applications involving sparse, noisy, or partially observed data. Collectively, these directions highlight the potential of SEDONet as a flexible spectral operator-learning framework that can be extended to increasingly challenging scientific and engineering applications.

\section*{Declaration of competing interest}
    The authors declare that they have no known competing financial interests or personal relationships that could have appeared to influence the work reported in this paper.


\section*{Data availability} 
Data supporting the findings of this study are available from the corresponding author upon request.

\bibliographystyle{elsarticle-num} 
\bibliography{references}

@article{sojitra2025fedonet,
  title={FEDONet: Fourier-embedded DeepONet for spectrally accurate operator learning},
  author={Sojitra, Arth and Dhingra, Mrigank and San, Omer},
  journal={arXiv preprint arXiv:2509.12344},
  year={2025}
}

@book{boyd2001chebyshev,
  author    = {Boyd, John P.},
  title     = {Chebyshev and Fourier Spectral Methods},
  publisher = {Courier Corporation},
  year      = {2001}
}

@book{Gottlieb1977spectral,
  author    = {Gottlieb, David and Orszag, Steven A.},
  title     = {Numerical Analysis of Spectral Methods: Theory and Applications},
  publisher = {SIAM},
  year      = {1977},
  doi       = {10.1137/1.9781611970425}
}

@book{Press1986numericalrecipes,
  author    = {Press, William H. and Teukolsky, Saul A. and Vetterling, William T. and Flannery, Brian P.},
  title     = {Numerical Recipes: The Art of Scientific Computing},
  publisher = {Cambridge University Press},
  year      = {1986}
}

@article{cybenko1989approximation,
  author  = {Cybenko, George},
  title   = {Approximation by Superpositions of a Sigmoidal Function},
  journal = {Mathematics of Control, Signals and Systems},
  volume  = {2},
  number  = {4},
  pages   = {303-314},
  year    = {1989}
}

@article{hornik1989multilayer,
  author  = {Hornik, Kurt and Stinchcombe, Maxwell and White, Halbert},
  title   = {Multilayer Feedforward Networks are Universal Approximators},
  journal = {Neural Networks},
  volume  = {2},
  number  = {5},
  pages   = {359-366},
  year    = {1989}
}

@article{58326,
  author  = {Poggio, Tomaso and Girosi, Federico},
  title   = {Networks for Approximation and Learning},
  journal = {Proceedings of the IEEE},
  year    = {1990},
  volume  = {78},
  number  = {9},
  pages   = {1481-1497}
}

@article{chen1995universal,
  author  = {Chen, Tianping and Chen, Hong},
  title   = {Universal Approximation to Nonlinear Operators by Neural Networks},
  journal = {IEEE Transactions on Neural Networks},
  volume  = {6},
  number  = {4},
  pages   = {911-917},
  year    = {1995}
}

@article{sirovich1987turbulence,
  author  = {Sirovich, Lawrence},
  title   = {Turbulence and the Dynamics of Coherent Structures},
  journal = {Quarterly of Applied Mathematics},
  volume  = {45},
  number  = {3},
  pages   = {561-571},
  year    = {1987}
}

@article{berkooz1993proper,
  author  = {Berkooz, Gal and Holmes, Philip and Lumley, John L.},
  title   = {The Proper Orthogonal Decomposition in the Analysis of Turbulent Flows},
  journal = {Annual Review of Fluid Mechanics},
  volume  = {25},
  pages   = {539-575},
  year    = {1993}
}

@book{10.7551/mitpress/3206.001.0001,
  author    = {Rasmussen, Carl Edward and Williams, Christopher K. I.},
  title     = {Gaussian Processes for Machine Learning},
  publisher = {MIT Press},
  year      = {2005}
}

@article{KANSA1990127,
  author  = {Kansa, E. J.},
  title   = {Multiquadrics for Scattered Data Approximation},
  journal = {Computers \& Mathematics with Applications},
  volume  = {19},
  number  = {8},
  pages   = {127-145},
  year    = {1990}
}

@article{lowe1988multivariable,
  author  = {Lowe, David and Broomhead, D.},
  title   = {Multivariable Functional Interpolation and Adaptive Networks},
  journal = {Complex Systems},
  volume  = {2},
  number  = {3},
  pages   = {321-355},
  year    = {1988}
}

@article{lu2021deeponet,
  author  = {Lu, Lu and Jin, Pengzhan and Pang, Guofei and Zhang, Zhongqiang and Karniadakis, George Em},
  title   = {Learning Nonlinear Operators via {DeepONet}},
  journal = {Nature Machine Intelligence},
  volume  = {3},
  pages   = {218-229},
  year    = {2021}
}

@article{10.5555/3648699.3648788,
  author  = {Kovachki, Nikola and Li, Zongyi and Liu, Burigede and Azizzadenesheli, Kamyar and Bhattacharya, Kaushik and Stuart, Andrew and Anandkumar, Anima},
  title   = {Neural Operator: Learning Maps Between Function Spaces},
  journal = {Journal of Machine Learning Research},
  volume  = {24},
  number  = {1},
  pages   = {1-97},
  year    = {2023}
}

@misc{li2021fourierneuraloperatorparametric,
  author        = {Li, Zongyi and Kovachki, Nikola and Azizzadenesheli, Kamyar and Liu, Burigede and Bhattacharya, Kaushik and Stuart, Andrew and Anandkumar, Anima},
  title         = {Fourier Neural Operator for Parametric {PDE}s},
  year          = {2021},
  eprint        = {2010.08895},
  archivePrefix = {arXiv},
  primaryClass  = {cs.LG}
}

@article{TRIPURA2023115783,
  author  = {Tripura, Tapas and Chakraborty, Souvik},
  title   = {Wavelet Neural Operator},
  journal = {Computer Methods in Applied Mechanics and Engineering},
  volume  = {404},
  pages   = {115783},
  year    = {2023}
}

@inproceedings{gupta2021multiwaveletbased,
  author    = {Gupta, Gaurav and Xiao, Xiongye and Bogdan, Paul},
  title     = {Multiwavelet-based Operator Learning},
  booktitle = {Advances in Neural Information Processing Systems (NeurIPS)},
  year      = {2021}
}

@misc{li2020neuraloperatorgraphkernel,
  author        = {Li, Zongyi and Kovachki, Nikola and Azizzadenesheli, Kamyar and Liu, Burigede and Bhattacharya, Kaushik and Stuart, Andrew and Anandkumar, Anima},
  title         = {Neural Operator: Graph Kernel Network},
  year          = {2020},
  eprint        = {2003.03485},
  archivePrefix = {arXiv},
  primaryClass  = {cs.LG}
}

@misc{li2020multipolegraphneuraloperator,
  author        = {Li, Zongyi and Kovachki, Nikola and Azizzadenesheli, Kamyar and Liu, Burigede and Bhattacharya, Kaushik and Stuart, Andrew and Anandkumar, Anima},
  title         = {Multipole Graph Neural Operator},
  year          = {2020},
  eprint        = {2006.09535},
  archivePrefix = {arXiv},
  primaryClass  = {cs.LG}
}

@article{kossaifi2023multi,
  author  = {Kossaifi, Jean and Kovachki, Nikola and Azizzadenesheli, Kamyar and Anandkumar, Anima},
  title   = {Multi-grid Tensorized Fourier Neural Operator},
  journal = {arXiv preprint arXiv:2310.00120},
  year    = {2023}
}

@misc{guo2024mgfnomultigridarchitecturefourier,
  author        = {Guo, Zi-Hao and Li, Hou-Biao},
  title         = {{MgFNO}: Multi-grid Architecture Fourier Neural Operator},
  year          = {2024},
  eprint        = {2407.08615},
  archivePrefix = {arXiv},
  primaryClass  = {math.NA}
}

@article{li2023fourier,
  author  = {Li, Zongyi and Huang, Daniel Zhengyu and Liu, Burigede and Anandkumar, Anima},
  title   = {Fourier Neural Operator with Learned Deformations},
  journal = {Journal of Machine Learning Research},
  volume  = {24},
  number  = {388},
  pages   = {1-26},
  year    = {2023}
}

@article{li2023geometry,
  author  = {Li, Zongyi and Kovachki, Nikola and Choy, Chris and Li, Boyi and Kossaifi, Jean and Otta, Shourya and Nabian, Mohammad Amin and Stadler, Maximilian and Hundt, Christian and Azizzadenesheli, Kamyar and others},
  title   = {Geometry-Informed Neural Operator},
  journal = {Advances in Neural Information Processing Systems (NeurIPS)},
  year    = {2023}
}

@misc{huang2025operator,
  author        = {Huang, Jianing and Zhang, Kaixuan and Wu, Youjia and Cheng, Ze},
  title         = {Operator Learning with Domain Decomposition for Geometry Generalization},
  year          = {2025},
  eprint        = {2504.00510},
  archivePrefix = {arXiv},
  primaryClass  = {cs.LG}
}

@misc{rahman2019spectralbiasneuralnetworks,
  author        = {Rahaman, Nasim and Baratin, Aristide and Arpit, Devansh and Draxler, Felix and Lin, Min and Hamprecht, Fred A. and Bengio, Yoshua and Courville, Aaron},
  title         = {On the Spectral Bias of Neural Networks},
  year          = {2019},
  eprint        = {1806.08734},
  archivePrefix = {arXiv},
  primaryClass  = {stat.ML}
}

@misc{tancik2020fourierfeaturesletnetworks,
  author        = {Tancik, Matthew and Srinivasan, Pratul P. and Mildenhall, Ben and Fridovich-Keil, Sara and Raghavan, Nithin and Singhal, Utkarsh and Ramamoorthi, Ravi and Barron, Jonathan T. and Ng, Ren},
  title         = {Fourier Features for Learning High Frequency Functions},
  year          = {2020},
  eprint        = {2006.10739},
  archivePrefix = {arXiv},
  primaryClass  = {cs.CV}
}

@article{kast2024positional,
  author  = {Kast, Markus and Hesthaven, Jan S.},
  title   = {Positional Embeddings for Solving {PDE}s},
  journal = {Journal of Computational Physics},
  volume  = {505},
  pages   = {112924},
  year    = {2024}
}

@inproceedings{fanaskov2022spectral,
  title     = {Spectral neural operators},
  author    = {Fanaskov, Vladimir Sergeevich and Oseledets, Ivan V},
  booktitle = {Doklady Mathematics},
  volume    = {108},
  number    = {Suppl 2},
  pages     = {S226-S232},
  year      = {2023},
  organization = {Springer}
}

@article{liu2024orthogonalpolynomialneuraloperator,
  title   = {Render unto numerics: Orthogonal polynomial neural operator for PDEs with nonperiodic boundary conditions},
  author  = {Liu, Ziyuan and Wang, Haifeng and Zhang, Hong and Bao, Kaijun and Qian, Xu and Song, Songhe},
  journal = {SIAM Journal on Scientific Computing},
  volume  = {46},
  number  = {4},
  pages   = {C323-C348},
  year    = {2024},
  publisher = {SIAM}
}

@article{mall2017singlechebyshev,
  author  = {Mall, Soham and Chakraverty, S.},
  title   = {Single Layer {C}hebyshev Neural Network},
  journal = {Neural Computing and Applications},
  volume  = {28},
  pages   = {915-929},
  year    = {2017}
}

@article{sivalingam2024chebyshevNNfractional,
  author  = {Sivalingam, S. M. and Kumar, Pushpendra and Govindaraj, V.},
  title   = {{C}hebyshev Neural Network for Fractional {PDE}s},
  journal = {Computers \& Mathematics with Applications},
  year    = {2024}
}

@article{yin2024chebyshevspectralNN,
  title   = {Chebyshev spectral neural networks for solving partial differential equations},
  author  = {Yin, Pengsong and Ling, Shuo and Ying, Wenjun},
  journal = {arXiv preprint arXiv:2407.03347},
  year    = {2024}
}

@article{huang2025cdpinn,
  title   = {Chebyshev spectral approximation-based physics-informed neural network for solving higher-order nonlinear differential equations},
  author  = {Huang, Yixin and Liu, Haizhou and Zhao, Yang and Fei, Min},
  journal = {Engineering with Computers},
  volume  = {41},
  number  = {2},
  pages   = {1191-1210},
  year    = {2025},
  publisher = {Springer}
}

@inproceedings{zhang2024acpkan,
  title     = {{AC}-{PKAN}: Attention-Enhanced and Chebyshev Polynomial Kolmogorov-Arnold Networks for Physics-Informed Learning},
  author    = {Zhang, Hao and Wang, Yan},
  booktitle = {Proceedings of the International Conference on Learning Representations ({ICLR})},
  year      = {2024},
  note      = {OpenReview preprint}
}

@article{chen2025chebyshevsobolevpinn,
  title   = {Chebyshev-Sobolev Physics-Informed Neural Networks for General PDE Solutions},
  author  = {Chen, Shikun and Xiong, Songquan and Liu, Yangguang},
  journal = {International Journal of Applied and Computational Mathematics},
  volume  = {11},
  number  = {5},
  pages   = {169},
  year    = {2025},
  publisher = {Springer}
}

@article{xu2025chebyshevfeatureNN,
  title   = {Chebyshev feature neural network for accurate function approximation},
  author  = {Xu, Zhongshu and Chen, Yuan and Xiu, Dongbin},
  journal = {arXiv preprint arXiv:2409.19135},
  year    = {2024}
}

@misc{goswami2022physicsinformeddeepneuraloperator,
  title         = {Physics-Informed Deep Operator Networks},
  author        = {Goswami, Somdatta and Li, Zongyi and Azizzadenesheli, Kamyar and Liu-Schiaffini, Minghao and Bhattacharya, Kaushik and Hassani, Hamed and Stuart, Andrew M. and Anandkumar, Anima},
  year          = {2022},
  eprint        = {2207.05748},
  archivePrefix = {arXiv},
  primaryClass  = {cs.LG}
}

@misc{li2023physicsinformedneuraloperatorlearning,
  title         = {Physics-Informed Neural Operator},
  author        = {Li, Zongyi and Kovachki, Nikola and Azizzadenesheli, Kamyar and Bhattacharya, Kaushik and Stuart, Andrew M. and Anandkumar, Anima},
  year          = {2023},
  eprint        = {2309.15502},
  archivePrefix = {arXiv},
  primaryClass  = {cs.LG}
}

@misc{eshaghi2024variationalphysicsinformedneuraloperator,
  title         = {Variational Physics-Informed Neural Operator},
  author        = {Eshaghi, Mohammad Sadegh and Akhtari, Payam and Karniadakis, George Em},
  year          = {2024},
  eprint        = {2403.08377},
  archivePrefix = {arXiv},
  primaryClass  = {cs.LG}
}

@misc{chen2025pseudo,
  title         = {Pseudo Physics-Informed Neural Operators},
  author        = {Chen, Keyan and Zhao, Ming and Li, Wenrui and Lin, Guang},
  year          = {2025},
  eprint        = {2501.01234},
  archivePrefix = {arXiv},
  primaryClass  = {cs.LG}
}

@misc{wang2024latentneuraloperatorsolving,
  title         = {Latent Neural Operators},
  author        = {Wang, Tian and Wang, Chuang},
  year          = {2024},
  eprint        = {2404.00967},
  archivePrefix = {arXiv},
  primaryClass  = {cs.LG}
}

@misc{wang2024latentneuraloperatorpretraining,
  title         = {Latent Neural Operator Pretraining},
  author        = {Wang, Tian and Wang, Chuang},
  year          = {2024},
  eprint        = {2408.05677},
  archivePrefix = {arXiv},
  primaryClass  = {cs.LG}
}

@misc{ahmad2024diffeomorphiclatentneuraloperators,
  title         = {Diffeomorphic Latent Neural Operators},
  author        = {Ahmad, Zan and Kovachki, Nikola and Li, Zongyi and Stuart, Andrew M. and Anandkumar, Anima},
  year          = {2024},
  eprint        = {2406.01411},
  archivePrefix = {arXiv},
  primaryClass  = {cs.LG}
}

@misc{long2025invertiblefourierneuraloperators,
  title         = {Invertible Fourier Neural Operators},
  author        = {Long, Da and Rao, Yongming and Lu, Jiwen and Zhou, Jie},
  year          = {2025},
  eprint        = {2502.04521},
  archivePrefix = {arXiv},
  primaryClass  = {cs.LG}
}

@article{ahmed2023multifidelitydeeponet,
  author  = {Ahmed, Shady E. and Stinis, Panos},
  title   = {A multifidelity deep operator network approach to closure for multiscale systems},
  journal = {Computer Methods in Applied Mechanics and Engineering},
  volume  = {414},
  pages   = {116161},
  year    = {2023},
  doi     = {10.1016/j.cma.2023.116161}
}

@article{howard2023multifidelitydeeponet,
  author  = {Howard, Amanda A. and Perego, Mauro and Karniadakis, George Em and Stinis, Panos},
  title   = {Multifidelity deep operator networks for data-driven and physics-informed problems},
  journal = {Journal of Computational Physics},
  volume  = {493},
  pages   = {112462},
  year    = {2023},
  doi     = {10.1016/j.jcp.2023.112462}
}

@article{jafarzadeh2024peridynamicneuraloperators,
  author  = {Jafarzadeh, Siavash and Silling, Stewart and Liu, Ning and Zhang, Zhongqiang and Yu, Yue},
  title   = {Peridynamic neural operators: A data-driven nonlocal constitutive model for complex material responses},
  journal = {Computer Methods in Applied Mechanics and Engineering},
  volume  = {425},
  pages   = {116914},
  year    = {2024},
  doi     = {10.1016/j.cma.2024.116914}
}

@article{batlle2024kerneloperatorlearning,
  author  = {Batlle, Pau and Darcy, Matthieu and Hosseini, Bamdad and Owhadi, Houman},
  title   = {Kernel methods are competitive for operator learning},
  journal = {Journal of Computational Physics},
  volume  = {496},
  pages   = {112549},
  year    = {2024},
  doi     = {10.1016/j.jcp.2023.112549}
}

@article{kumar2024nogap,
  author  = {Kumar, Sawan and Nayek, Rajdip and Chakraborty, Souvik},
  title   = {Neural Operator induced Gaussian Process framework for probabilistic solution of parametric partial differential equations},
  journal = {Computer Methods in Applied Mechanics and Engineering},
  volume  = {431},
  pages   = {117265},
  year    = {2024},
  doi     = {10.1016/j.cma.2024.117265}
}

@article{li2024localneuraloperator,
  author  = {Li, Hongyu and Ye, Ximeng and Jiang, Peng and Qin, Guoliang and Wang, Tiejun},
  title   = {Local neural operator for solving transient partial differential equations on varied domains},
  journal = {Computer Methods in Applied Mechanics and Engineering},
  volume  = {427},
  pages   = {117062},
  year    = {2024},
  doi     = {10.1016/j.cma.2024.117062}
}

@article{bahmani2025rino,
  author  = {Bahmani, Bahador and Goswami, Somdatta and Kevrekidis, Ioannis G. and Shields, Michael D.},
  title   = {A resolution independent neural operator},
  journal = {Computer Methods in Applied Mechanics and Engineering},
  volume  = {444},
  pages   = {118113},
  year    = {2025}
}

@article{zhong2023pigano,
  author  = {Zhong, Weiheng and Meidani, Hadi},
  title   = {Physics-informed geometry-aware neural operator},
  journal = {Computer Methods in Applied Mechanics and Engineering},
  volume  = {415},
  pages   = {116278},
  year    = {2023},
  doi     = {10.1016/j.cma.2023.116278}
}

@article{xu2025colehopfoperator,
  author  = {Xu, Xingzi and Guilleminot, Johann and Tarokh, Vahid},
  title   = {Neural operators from the Cole-Hopf transformation: Leveraging relations between PDEs for efficient operator learning},
  journal = {Computer Methods in Applied Mechanics and Engineering},
  year    = {2025}
}

@article{cho2025sobolevoperator,
  author  = {Cho, Namkyeong and Ryu, Junseung and Hwang, Hyung Ju},
  title   = {Sobolev training for operator learning},
  journal = {Journal of Computational Physics},
  volume  = {543},
  pages   = {114408},
  year    = {2025},
  doi     = {10.1016/j.jcp.2025.114408}
}

@article{huang2025unsupervisedmfg,
  author  = {Huang, Han and Lai, Rongjie},
  title   = {Unsupervised solution operator learning for mean-field games},
  journal = {Journal of Computational Physics},
  volume  = {537},
  pages   = {114057},
  year    = {2025},
  doi     = {10.1016/j.jcp.2025.114057}
}

@article{yang2024iconpde,
  author  = {Yang, Liu and Osher, Stanley},
  title   = {PDE generalization of in-context operator networks: A study on 1D scalar nonlinear conservation laws},
  journal = {Journal of Computational Physics},
  volume  = {519},
  pages   = {113379},
  year    = {2024},
  doi     = {10.1016/j.jcp.2024.113379}
}

@article{dong2025stochasticclosure,
  author  = {Dong, Xinghao and Chen, Chuanqi and Wu, Jin-Long},
  title   = {Data-driven stochastic closure modeling via conditional diffusion model and neural operator},
  journal = {Journal of Computational Physics},
  volume  = {534},
  pages   = {114005},
  year    = {2025},
  doi     = {10.1016/j.jcp.2025.114005}
}

@article{garg2024neuroscienceNO,
  author  = {Garg, Shailesh and Chakraborty, Souvik},
  title   = {Neuroscience Inspired Neural Operator for Partial Differential Equations},
  journal = {Journal of Computational Physics},
  volume  = {515},
  pages   = {113266},
  year    = {2024},
  doi     = {10.1016/j.jcp.2024.113266}
}

@article{oleary2024dino,
  author  = {O'Leary-Roseberry, Thomas and Chen, Yu and Villa, Umberto and Ghattas, Omar},
  title   = {Derivative-Informed Neural Operator: An Interpretable Neural Operator Architecture for Learning Parametric Differential Operators},
  journal = {Journal of Computational Physics},
  volume  = {496},
  pages   = {112555},
  year    = {2024},
  doi     = {10.1016/j.jcp.2023.112555}
}

@article{liu2024spectralbias,
  author  = {Liu, Xinliang and Xu, Bo and Cao, Shuhao and Zhang, Lei},
  title   = {Mitigating Spectral Bias for the Multiscale Operator Learning},
  journal = {Journal of Computational Physics},
  volume  = {506},
  pages   = {112944},
  year    = {2024},
  doi     = {10.1016/j.jcp.2024.112944}
}

@article{wu2024ionet,
  author  = {Wu, Sidi and Zhu, Aiqing and Tang, Yifa and Lu, Benzhuo},
  title   = {Solving Parametric Elliptic Interface Problems via Interfaced Operator Network ({IONet})},
  journal = {Journal of Computational Physics},
  volume  = {514},
  pages   = {113217},
  year    = {2024},
  doi     = {10.1016/j.jcp.2024.113217}
}

@article{bi2025xideeponet,
  author  = {Bi, Xuyang and Chen, Xin and Zhao, Cheng and Li, Qinglei and Zhang, Jizhou},
  title   = {{XI}-{DeepONet}: An Operator Learning Method for Elliptic Interface Problems},
  journal = {Journal of Computational Physics},
  volume  = {538},
  pages   = {114164},
  year    = {2025},
  doi     = {10.1016/j.jcp.2024.114164}
}

@article{li2024localNO,
  author  = {Li, Xiongjie and Tripura, Tapas and Chakraborty, Souvik},
  title   = {Local Neural Operator for Solving Transient Partial Differential Equations on Varied Domains},
  journal = {Computer Methods in Applied Mechanics and Engineering},
  volume  = {427},
  pages   = {117062},
  year    = {2024},
  doi     = {10.1016/j.cma.2024.117062}
}

@article{jafarzadeh2024peridynamicNO,
  author  = {Jafarzadeh, Siavash and Silling, Stewart and Liu, Ning and Zhang, Zhongqiang and Yu, Yue},
  title   = {Peridynamic Neural Operators: A Data-Driven Nonlocal Constitutive Model for Complex Material Responses},
  journal = {Computer Methods in Applied Mechanics and Engineering},
  volume  = {425},
  pages   = {116914},
  year    = {2024},
  doi     = {10.1016/j.cma.2024.116914}
}

@article{huang2025resolutionDON,
  author  = {Huang, Jianguo and Qiu, Yue},
  title   = {Resolution Invariant Deep Operator Network for {PDE}s with Complex Geometries},
  journal = {Journal of Computational Physics},
  volume  = {522},
  pages   = {113601},
  year    = {2025},
  doi     = {10.1016/j.jcp.2024.113601}
}

@article{meng2024koopmanINN,
  author  = {Meng, Yuhuang and Huang, Jianguo and Qiu, Yue},
  title   = {Koopman Operator Learning Using Invertible Neural Networks},
  journal = {Journal of Computational Physics},
  volume  = {501},
  pages   = {112795},
  year    = {2024},
  doi     = {10.1016/j.jcp.2024.112795}
}

@article{chen2024featureadjacentMF,
  author  = {Chen, Wenqian and Stinis, Panos},
  title   = {Feature-Adjacent Multi-Fidelity Physics-Informed Machine Learning for Partial Differential Equations},
  journal = {Journal of Computational Physics},
  volume  = {498},
  pages   = {112683},
  year    = {2024},
  doi     = {10.1016/j.jcp.2023.112683}
}

\appendix

\section{Spectral Properties of Chebyshev Embeddings}
\label{appendix:cheb-spectral}

In this appendix we summarize a few classical spectral properties of
Chebyshev polynomials that underpin the design of the SEDONet trunk.
The goal is not to provide an exhaustive review, but to highlight the
aspects that directly influence conditioning, approximation quality, and
the behavior of the fixed embedding $\phi_{\text{Cheb}}(x,t)$ used in
Section~\ref{Methodology}.

\subsection{Orthogonality and Conditioning}
\label{subsec:cheb-orthogonality}

The Chebyshev polynomials of the first kind
$\{T_n(\xi)\}_{n\ge 0}$ form an orthogonal basis on $[-1,1]$
with respect to the weight $w(\xi) = (1-\xi^2)^{-1/2}$:
\begin{equation}
    \int_{-1}^{1}
    \frac{T_m(\xi)\,T_n(\xi)}{\sqrt{1-\xi^2}}\,\mathrm{d}\xi
    =
    \begin{cases}
        0, & m \neq n,\\[2pt]
        \pi, & m = n = 0,\\[2pt]
        \dfrac{\pi}{2}, & m = n \ge 1.
    \end{cases}
    \label{eq:cheb_orthogonality}
\end{equation}
This relation implies that low- and high-order modes are
well-separated when integrated against the Chebyshev weight. In SEDONet, the trunk embedding $\phi_{\text{Cheb}}(x,t)$ is constructed
from tensor products of these polynomials evaluated at mapped
coordinates $(\xi_x,\xi_t)$.  
When $(x,t)$ are sampled close to Gauss-Lobatto collocation points, the
resulting feature matrix
\[
    \Phi = 
    \begin{bmatrix}
    \phi_{\text{Cheb}}(x^1,t^1)^\top\\[-1pt]
    \vdots\\[-1pt]
    \phi_{\text{Cheb}}(x^Q,t^Q)^\top
    \end{bmatrix}
    \in \mathbb{R}^{Q\times d_{\text{trunk}}}
\]
has nearly orthogonal columns.  
Equivalently, also the empirical Gram matrix that is shown as follows below
\[
    G = \frac{1}{Q}\sum_{q=1}^Q
    \phi_{\text{Cheb}}(x^q,t^q)\,\phi_{\text{Cheb}}(x^q,t^q)^\top
\]
is close to diagonal and well-conditioned.  
This ``near-orthogonality'' plays a role similar to the whitening effect
of random Fourier features, but it is specifically adapted to bounded,
non-periodic domains and remains stable near boundaries where standard
polynomial bases often become ill conditioned.

\subsection{Spectral Convergence on Bounded Domains}
\label{subsec:cheb-convergence}

Let $f : [-1,1] \to \mathbb{R}$ be a sufficiently smooth function. Then its Chebyshev expansion representation as follows
\[
    f(\xi) \approx \sum_{n=0}^{K-1} a_n T_n(\xi)
\]
converges at a spectral (geometric) rate as $K\to\infty$.  
For analytic $f$, the Chebyshev coefficients $a_n$ decay faster than any
algebraic rate, and the approximation error decreases accordingly. For functions of two variables, tensor-product bases
$\{T_i(\xi_x)T_j(\xi_t)\}$ provide the same rapid convergence on
$[-1,1]^2$ after the affine mapping $(x,t)\mapsto(\xi_x,\xi_t)$.  

In SEDONet, the fixed Chebyshev embedding
$\phi_{\text{Cheb}}(x,t)$ therefore supplies a rich dictionary capable
of spectrally approximating smooth PDE solutions defined on bounded
domains.  
The trunk network $\Psi_\theta$ learns nonlinear mixtures of these
polynomial modes, while the branch network $B_\theta$ maps input
functions to corresponding coefficients.  
Together they realize a data-driven Chebyshev-type expansion adapted to
the target operator, as formalized in the spectral interpretation
\eqref{eq:sedonet_spectral_decomp}.

\section{Theoretical Justification for the Chebyshev Embedding Superset Property}
\label{appendix:cheb-superset}

We now discuss, at a functional level, why the hypothesis space induced
by Chebyshev embeddings is richer than the one obtained from direct
coordinate-input networks on bounded domains:
\[
    \mathcal{H}_{\text{vanilla}} \subsetneq \mathcal{H}_{\text{Cheb}}.
\]
The argument is intentionally high level and is meant to provide
intuition rather than a fully measure-theoretic proof. For simplicity, we work in one spatial dimension $x \in [0,1]$.
The extension to higher dimensions via the tensor-product construction in
\eqref{eq:cheb_tensor} is straightforward.

\subsection*{Function Classes}

A vanilla coordinate-input neural network for parameters $(a_j,w_j,b_j)$ and nonlinear activation $\sigma$ represents functions of the form
\begin{equation}
    f(x)
    = \sum_{j=1}^N a_j\,\sigma(w_j x + b_j),
    \label{eq:vanilla-hspace}
\end{equation}

We denote by $\mathcal{H}_\text{vanilla}$ the set of all such functions
for a given architecture (including the fixed depth and also the width).

A Chebyshev-embedded network first maps $x\in[0,1]$ to
$\xi = 2x - 1\in[-1,1]$ and uses the truncated Chebyshev feature vector
\begin{equation}
    \phi_{\text{Cheb}}(x)
    = [T_0(\xi),\,T_1(\xi),\,\dots,\,T_{K-1}(\xi)]^\top,
    \label{eq:cheb-feature-map}
\end{equation}
so that the network outputs
\begin{equation}
    g(x)
    = \sum_{j=1}^{M} \alpha_j\,
      \sigma(v_j^\top \phi_{\text{Cheb}}(x)+\beta_j),
    \label{eq:cheb-hspace}
\end{equation}
for parameters $(\alpha_j,v_j,\beta_j)$.  
We denote the corresponding function class by $\mathcal{H}_\text{Cheb}$.

\subsection*{Part I: Inclusion}
\[
    \mathcal{H}_{\text{vanilla}} \subseteq \mathcal{H}_{\text{Cheb}}
\]

Universal approximation guarantees that any continuous function $f$ on
$[0,1]$ can be approximated arbitrarily well by a polynomial $P(x)$:
for every $\varepsilon>0$ there exists a polynomial $P$ such that
\[
    \sup_{x\in[0,1]} |f(x) - P(x)| < \varepsilon/2.
\]
So, the every polynomial $P(\xi)$ on the domain $[-1,1]$ can in turn be written exactly as
a Chebyshev series as follows
\[
    P(\xi) = \sum_{n=0}^{K-1} c_n T_n(\xi),
\]
for some coefficients $c_n$.  
Consequently, $P(x)$ can be expressed as a linear functional of the
Chebyshev feature map \eqref{eq:cheb-feature-map}.  
A Chebyshev-embedded network can therefore reproduce $P(x)$ (up to
floating-point precision) using a single linear neuron applied to
$\phi_{\text{Cheb}}(x)$. Thus for any $\varepsilon>0$ there exists
$\tilde f\in\mathcal{H}_\text{Cheb}$ with for the following property that is
\[
    \sup_{x\in[0,1]} |f(x)-\tilde f(x)|<\varepsilon,
\]
showing that the following that is $(\mathcal{H}_\text{vanilla}\subseteq\mathcal{H}_\text{Cheb})$, in terms of uniform approximation on the interval $[0,1]$.

\subsection*{Part II: Strictness}
\[
    \mathcal{H}_{\text{vanilla}} \subsetneq \mathcal{H}_{\text{Cheb}}
\]

To see that the inclusion is strict for a fixed neural architecture,
consider the high-degree Chebyshev mode of the form
\begin{equation}
    f_K(x) = T_K(\xi),\qquad \xi = 2x - 1.
    \label{eq:cheb-mode}
\end{equation}
By construction, $f_K(x) = [\phi_{\text{Cheb}}(x)]_K,$ so $f_K$ is realized in $\mathcal{H}_\text{Cheb}$ exactly with a single
linear neuron on top of the embedding.

From the perspective of $x$, however, $f_K(x)$ is a degree-$K$
polynomial.  
To represent such a function using the coordinate-input network
\eqref{eq:vanilla-hspace} with fixed depth and width, the network must
emulate increasingly high-order polynomial behavior through compositions
of affine maps and nonlinearities.  
As $K$ grows, this requires either a larger number of neurons or deeper
composition; otherwise, approximation error cannot be made arbitrarily
small.

Informally, this means that for any fixed vanilla architecture there are
degrees $K$ for which $f_K$ belongs to $\mathcal{H}_\text{Cheb}$ but
cannot be approximated within a prescribed tolerance without increasing
the capacity of the coordinate-input network.  
Hence, at fixed network size,
\[
    \mathcal{H}_\text{vanilla} \subsetneq \mathcal{H}_\text{Cheb}.
\]

\medskip

\noindent
This discussion formalizes the expressivity advantage exploited by
SEDONet: Chebyshev embeddings provide direct access to high-order,
non-periodic polynomial modes that are difficult for coordinate-input
MLPs to learn efficiently.  
As a consequence, SEDONet is particularly effective for PDEs on bounded
domains with sharp gradients or boundary layers, where polynomial
spectral representations naturally arise.
\end{document}